\documentclass{tlp}

\usepackage{amsmath}
\usepackage{graphicx}
\usepackage{multirow}
\usepackage{amsfonts}
\usepackage{booktabs}
\usepackage{color}
\usepackage{listings}
\usepackage{multirow}
\usepackage{subcaption}
\usepackage{tabularx}

\usepackage{lineno,hyperref}
\modulolinenumbers[5]

\newcounter{example}[section]
\newenvironment{example}[1][]{\refstepcounter{example}\par\medskip
   \noindent \textbf{Example~\theexample. #1} \rmfamily}{\medskip}
   
\def\punto{\hspace*{\fill}\Box}

\begin{document}

\lefttitle{Cambridge Author}

\jnlPage{1}{8}
\jnlDoiYr{2021}
\doival{10.1017/xxxxx}

\title[Extended High-Utility Pattern Mining with ASP]{Extended High Utility Pattern Mining: An Answer Set Programming Based Framework and Applications}

\begin{authgrp}
\lefttitle{F. Cauteruccio and G. Terracina}
\author{\sn{Francesco} \gn{Cauteruccio}}
\affiliation{DII, Polytechnic University of Marche, Ancona (Italy) \\ \texttt{f.cauteruccio@univpm.it}}
\author{\sn{Giorgio} \gn{Terracina}}
\affiliation{DEMACS, University of Calabria, Rende (Italy) \\ \texttt{terracina@mat.unical.it}}
\end{authgrp}

\history{\sub{xx xx xxxx;} \rev{xx xx xxxx;} \acc{xx xx xxxx}}

\maketitle

\begin{abstract}

 Detecting sets of relevant patterns from a given dataset is an important challenge in data mining. The relevance of a pattern, also called utility in the literature, is a subjective measure and can be actually assessed from very different points of view. Rule-based languages like Answer Set Programming (ASP) seem well suited for specifying user-provided criteria to assess pattern utility in a form of constraints; moreover, declarativity of ASP allows for a very easy switch between several criteria in order to analyze the dataset from  different points of view. In this paper, we make steps toward extending the notion of High Utility Pattern Mining (HUPM); in particular we introduce a new framework that allows for new classes of utility criteria not considered in the previous literature. We also show how recent extensions of ASP with external functions can support a fast and effective encoding and testing of the new framework. To demonstrate the potential of the proposed framework, we exploit it as a building block for the definition of an innovative method for predicting ICU admission for COVID-19 patients. Finally, an extensive experimental activity demonstrates both from a quantitative and a qualitative point of view the effectiveness of the proposed approach. Under consideration in Theory and Practice of Logic Programming (TPLP).

\end{abstract}


\section{Introduction}
\label{sec:introduction}

Pattern mining is one of the data mining branches that attracted vast attention in the literature. Pattern mining algorithms extract human-understandable knowledge from databases and have fostered a growing interest in the development of scalable and flexible methods for data analysis. In this context, flexibility is intended as the ability to incorporate users' prior knowledge and  multiple criteria to measure the relevance of a pattern in the analysis process. These criteria can be modelled in the form of constraints to validate a set of candidate patterns. The first and most common studied constraint is the frequency threshold, where a pattern is validated only if it appears sufficiently often~\citep{AgSr94}. Frequent pattern mining is a fairly  well studied problem and effective solutions are available also in the logic programming arena \citep{Jarvisalo11, Gebser*16,GuMoQu14}. 

However, frequency alone may be of little interest in many cases. As an example, consider a sales database and the pattern $\{\text{\emph{windshield washer fluid}}, \text{\emph{new windshield wipers}}\}$; this pattern can be very frequent but uninteresting, as it represents a common purchase. But what if we also consider the price of the products and a constraint on the minimum profit given by the corresponding purchase pattern? In this case, the pattern $\{\text{\emph{car}}, \text{\emph{car alarm}}\}$ might be more interesting than the previous one, even if less frequent. 

In light of this last consideration, the academic community has begun to emphasise that pattern validity can be assessed according to \emph{utility functions}~\citep{Fou-Vi*19,Gan*19a}; the introduction of the notion of the utility of an item and, more generally, of a pattern made it possible to develop a new generation of pattern mining approaches called High Utility Pattern Mining - HUPM~\citep{Fou-Vi*19,Gan*19a,YaHaGe06}.

However, the basic assumption of HUPM is that each item is associated with \emph{one}, \emph{static}, external utility; this limits the flexibility expected from modern data analysis processes. For instance, continuing the example on the sales database, it would not be possible to devise validation constraints that dynamically combine price and minimum packaging size of the various products composing the pattern in the utility calculation; this could be very valuable, for example, in logistics optimization applications. Generalizing, it would be really important to be able to combine, in a flexible way, different aspects of the items, which we call \emph{facets} in the following,  to compute patterns' utility. 

Another limitation of HUPM lies in the fact that the expected input is a {\em flat} representation of transactions; this allows only local utility notions to be defined. On the contrary, a multi-layered representation of the data, coupled with the possibility of combining different facets in utility functions, may allow more advanced pattern constraints to be defined. As an example, by grouping purchases by customer and considering the degree of each customer  satisfaction as a facet, one can identify patterns with a good correlation between customer purchases and their satisfaction. This is not possible with classical HUPM methods.

Finally, it can be useful to incorporate users' prior knowledge into pattern constraints, in the form of \emph{pattern masks}, to assess pattern utility. For instance, it may be useful to state that only certain combinations of product categories are relevant for the analysis. Again, this is not currently possible in the classical HUPM setting. 

A first contribution of the present paper is the definition of a framework, which we call e-HUPM, that generalizes the classical HUPM in the following directions:  

\begin{itemize}
    \item \emph{Dataset representation}. A multi-layered representation of the input is defined, where aggregation levels of transactions,  objects, and containers are conventionally identified.
    \item \emph{Facets}. We introduce the notion of facet, which can be associated with an item, a transaction, an object or a container; each of these elements may be characterized by more than one facet.
    \item \emph{Utility functions}. In order to make the most of facets and layers, we introduce a taxonomy of utility functions classes, most of which are not allowed in classical HUPM.
   
    \item \emph{Pattern masks}. We introduce the notion of pattern mask, in order to specify structural or semantic constraints patterns must comply with.  
\end{itemize}

To solve the pattern mining problem introduced above, a guess-and-check resolution scheme, with the support of rule-based languages such as Answer Set Programming (ASP), seems to be quite intuitive and effective. 
In particular, we have seen that solutions for frequent pattern mining are already available in ASP \citep{Jarvisalo11, Gebser*16,GuMoQu14}; however, the constraints we are concerned with in this work to validate patterns come in the form of potentially complex functions, which are difficult, or even impossible, to express in ASP. In order to solve this issue, we resort to recent extensions of ASP systems, such as DLVHEX~\citep{Eiter*16,Eiter*18}, WASP \citep{DoRi20}, clingo~\citep{Gebser*19},  etc., which allow external calculation functions, usually written in Python, to be integrated into ASP programs.  

A second contribution of the present work is a modular ASP encoding of the proposed framework, so that, given a pool of alternative encodings for each module, even those unfamiliar with ASP can set up their own variants  in a way that provides a high degree of flexibility for data analysis. 

Finally, as a third contribution, we exploit the proposed framework and the corresponding ASP-based solution as a building block for the definition of an innovative method for predicting the ICU admission for COVID-19 patients, based on patients' blood results and vital signs.

The remainder of the paper is organized as follows.  In Section~\ref{sec:ehupm-framework} the general framework is proposed, along with all of its components and the problem definition. Section~\ref{sec:asp} is devoted to the design of the ASP solution. The application of our framework to a biomedical context is presented in Section~\ref{sec:application}, whereas an extensive experimental evaluation is presented in Section~\ref{sec:experiments}. In Section~\ref{sec:related-work} a broad overview of related work is presented. Finally, in Section~\ref{sec:conclusion} we draw our conclusions and highlight future research directions.

\section{A General Framework for Extending High Utility Pattern Mining (e-HUPM)}
\label{sec:ehupm-framework}
\label{sec:ehupm}
In this section we present our framework. First, we briefly recall the classical background definitions related to the HUPM problem. Then, we show how we extend the classical problem in several ways. In particular, we first extend the concept of \emph{Transaction database} with \emph{Containers} and \emph{Objects}; then we extend the concept of utility with the notion of \emph{facet}. After this, we introduce a new classification of \emph{pattern utility} functions and the notion of \emph{pattern mask}.  Finally, we provide a formal definition of the problem addressed in this work. 

\subsection{Background}
\label{sub:background}
We now briefly summarize the classical definitions and notations for the HUPM problem based on the ones presented in~\citep{Fou-Vi*19}. It is worth pointing out that several variants of this problem have been proposed in the literature; since it is out of the scope of the paper covering all of them here, we focus on the most classical one.

A quantitative transaction database $D$ is composed of a set of transactions and denoted as $D=\{T_1, T_2, \ldots, T_n\}$. Each transaction $T_p$ is uniquely identified by a transaction identifier ($tid_p$) and contains a set of items $I_{p}$ where each $I_{p} \subseteq I=\{i_1, i_2, \ldots, i_m\}$. Each item $i \in I$ is associated with an \emph{external utility} $eu(i)$ and every occurrence of $i$ in a transaction $T_p$ is associated with an \emph{internal utility} $iu(i,T_p)$ which generally represents the quantity of $i$ in $T_p$. In the classical definition of HUPM both external and internal utilities are positive numbers. 
The objective of HUPM is the identification of sets of items (patterns) that 
present a high utility, i.e., a utility higher than a certain threshold $th_u$.

The utility of an item $i$ in a transaction $T_p$ is obtained as $eu(i) \times iu(i,T_p)$. Given a pattern $P$ appearing in a transaction $T_p$, the utility of $P$ in $T_p$ is denoted as $tu(P,T_p)$ and is computed as $\sum_{i \in P} eu(i) \times iu(i,T_p)$.  

Given the set $T_{P}$ of transactions containing occurrences of the pattern $P$, the utility of $P$ in the database $D$ is denoted by $u(P)$ and computed as $u(P)=\sum_{T_p \in T_{P}} tu(P,T_p)$.

\paragraph{Problem definition.} Given a pattern $P$, $P$ is a high utility pattern if its utility $u(P)$ computed over its occurrences in $D$ is greater than a utility threshold $th_u$. 
The problem of high utility pattern mining is to discover all high utility patterns in a database $D$.

\subsection{Databases, containers and objects.}
The first extension we propose with respect to the classical HUPM relates to the database representation. In the classical context, a database is simply composed of a flat set of transactions. Here we assume that a database of transactions is organized in different abstraction layers. This allows us to include more sophisticated utility functions in order to analyze, as an example, correlations among different transactions at different abstraction levels. Here we propose a possible hierarchy for the database, which fixes some boundaries for the definition of the problem; clearly more general structures and more layers can be defined in different contexts. The proposed reference database structure is as follows:

\begin{center}
    \textit{Database} $\rightarrow$ \textit{Container} $\rightarrow$ \textit{Object} $\rightarrow$ \textit{Transaction}
\end{center}

In particular, given a database $D$ and a set of transactions $\{T_1, T_2, \ldots, T_n\}$, $D$ is organized as a set of \emph{containers} $C=\{C_1, C_2, \ldots, C_r \}$ where each container $C_s$ can be associated with a set of \emph{objects}  $O=\{O_1, O_2, \ldots, O_t\}$ and each object $O_u$ contains a set of transactions $\{T_1, T_2, \ldots, T_v\}$. Clearly each transaction is composed of a (possibly ordered) set of items. Each transaction belongs to precisely one object and each object is associated with precisely one container. 
Observe that this representation allows covering several interesting application scenarios. 

\begin{example}[\emph{(Running example)}]\label{example:1}
In order to better describe our framework, let us introduce a running example concerning scientific reviews, which will be explored further in the experiments. This context presents several peculiarities captured by our framework. In particular, we build upon the paper reviews use case presented in~\citep{ChGoMu20}; it deals with papers, reviews and annotated sentences.
Each paper is a container in our model; for each paper, several reviews are available: each review is an object in our model. Each review contains a list of sentences: each sentence is a transaction in our model. 
 
$\punto$
\end{example}

\subsection{Utilities and facets.} 
\label{sub:facets-utilities}
The second main generalization deals with utility representation. As a matter of fact, a great limit of the classical definition of HUPM, and its variants, is that each item is associated with a unique, external, and fixed value of utility. 
We next extend the notion of utility with the concept of \emph{facet}. In fact, in several contexts, the utility of an item can be defined from different perspectives, which we call facets. Then, each item can be associated with a list of values defining its utility from different perspectives. Moreover, given the new organization of the database, facets can be defined also for transactions, objects, and containers. Formally:

{\em Item utility vector.} Given an item $i$, the utility of $i$ is defined by an \emph{item utility vector} $IU_{i}=[iu_1, iu_2, \ldots, iu_l]$, where each $iu_k$ describes a certain facet of $i$.

{\em Transaction utility vector.} Given a transaction $T_p$, the utility of $T_p$ is defined by a \emph{transaction utility vector} $TU_{T_p}=[tu_1, tu_2, \ldots, tu_m]$, where each $tu_k$ describes a facet of $T_p$. Observe that these facets for the transaction describe properties of the transaction as a whole and represent a different information than the standard internal utility of an item $i$ in the transaction $T_p$. In order to keep the compatibility with the classical problem, we assume that the internal utility of $i$ in $T_p$ is available and represented as $q(i,T_p)$. 

{\em Object utility vector.} Given an object $O_u$, the utility of $O_u$ is defined by an \emph{object utility vector} $OU_{O_u}=[ou_1, ou_2, \ldots, ou_n]$, where each $ou_k$ describes a facet of $O_u$.

{\em Container utility vector.} Given a container $C_s$, the utility of $C_s$ is defined by a vector $CU_{C_s}=[cu_1, cu_2, \ldots, cu_o]$, where each $cu_k$ describes a facet of $C_s$.

It is worth observing that the length of any of the above utility vectors could be 0 if no interesting facet can be defined for that vector. The list of facets is fixed at problem modelling stage; however, we assume that all utility vectors of a certain type have the same number of facets. All utilities introduced above are not constrained to be numeric values. The interpretation and combination of utilities is left to the pattern utility computation function, introduced in the next section.

\begin{table}[t!]
	\centering
	\caption{Terminology and facets for the paper reviews use case running example introduced in Example~\ref{example:1}}
	\label{tab:papers-vs-hupm}
	{\footnotesize
	\begin{tabular}{llll}
    \topline    
      e-HUPM & Use case   & Facets & Domain of the facet
      \midline
      Database & Set of reviews  &  - & \\
      Container & Paper & (Decision) & (\{0 (Reject), 1 (Accept)\})\\
      Object & Review & (Rating, Confidence) & (\{1-10\}, \{1-5\}) \\
      Transaction & Sentence &  (Appropriateness, Clarity, Originality, &  \\
      & & Soundness, Comparison, Substance, & \{-1 (Negative), 1 (Positive), \\
      & & Impact, Recommendation) & 0 (Neutral or Absent) \}\\
      Item & Word & -
      \botline
	\end{tabular}
}
\end{table}

\begin{table}[t!]
\centering
\caption{Examples of container, object and transaction utility vectors for the paper reviews use case running example}
\label{tab:example-2}
{\footnotesize
\begin{tabular}{cl|cl}
    \topline
    Database elements & Utility vectors & Database elements & Utility vectors
    \midline
    \multicolumn{2}{l|}{Containers (papers)} & \multicolumn{2}{l}{Transaction (reviews' sentence)} \\
    $C_1$ & $CU_{C_1} = [0]$ &  $S_1$ & $TU_{S_1} = [-1,-1,\ \ 0,\ \ 1,\ \ 0,\ \ 0,\ \ 0,-1]$\\
    $C_2$ & $CU_{C_2} = [1]$ & $S_2$ & $TU_{S_2} = [\ \ 1,\ \ 0,\ \ 1,\ \ 0,-1,-1,\ \ 0,\ \ 0]$ \\ 
    \multicolumn{2}{l|}{Object (reviews)} & $S_3$ & $TU_{S_3} = [\ \ 1,\ \ 1,\ \ 1,\ \ 0,\ \ 0,\ \ 0,\ \ 0,\ \ 1]$ \\
    $R_1$ & $OU_{R_1} = [2, 4]$  & $S_4$ & $TU_{S_4} = [\ \ 0,\ \ 1,-1,\ \ 1,\ \ 1,\ \ 0,\ \ 1,\ \ 1]$\\
    $R_2$ & $OU_{R_2} = [4, 3]$ \\
    $R_3$ & $OU_{R_3} = [9, 3]$
    \botline
\end{tabular}
}
\end{table}

\begin{table}[t]
\centering
\caption{An excerpt of the transactions contained in the paper reviews use case, along with some objects and containers, used in the running example}
\label{tab:example-3}
{\footnotesize
\begin{tabular}{cccl}
    \topline
    Container & Object & TID & Transaction
    \midline
    \multirow{2}{*}{$C_1$} & $R_1$ & $S_1$ & $\{(\textsf{paper},2),(\textsf{hard},1),(\textsf{narrow},1)\}$ \\
    & $R_2$ & $S_2$ & $\{(\textsf{problem},1),(\textsf{paper},1),(\textsf{concern},1),(\textsf{reproducibility},1)\}$
    \midline
    \multirow{2}{*}{$C_2$} & \multirow{2}{*}{$R_3$} & $S_3$ & $\{(\textsf{paper},1),(\textsf{readable},1)\}$ \\
    & & $S_4$ & $\{(\textsf{paper},1),(\textsf{good},1),(\textsf{experiment},1),(\textsf{reproducibility},1)\}$
    \botline
\end{tabular}
}
\end{table}

\begin{example}[\emph{(Running example)}]\label{example:2}
Table~\ref{tab:papers-vs-hupm} illustrates the correspondence between our e-HUPM framework and the paper reviews use case adopted from~\citep{ChGoMu20}. The table depicts the selected facets for each domain element and its representation in the database, e.g., facets for containers, objects and transactions; the database and items have no associated facets. Each container represents a paper and for a container $C_s$ we have one single facet, that is the decision. Reviews information are represented by objects and for an object $O_u$ we have two facets, namely rating and confidence. Each review's sentence is a transaction and for a transaction $T_p$ we have the facets corresponding to the eight annotated aspects defined in~\citep{ChGoMu20} and reported in Table~\ref{tab:papers-vs-hupm}. Finally, each sentence's word is an item. Table~\ref{tab:example-2} depicts the utility vectors for container, object, and transaction respectively, relative to a simple instance excerpt adapted from the dataset; this excerpt is reported in Table~\ref{tab:example-3} for completeness of presentation.  
In this table, for a better readability, we report the word corresponding to each item. We point out that a pre-processing pipeline is applied to the sentences with usual stemming and lemmatization procedures. Note that each transaction $T_p$ is a set of pairs $(i, q(i, T_p))$ where $i$ is an item and $q(i, T_p)$ its quantity in the transaction $T_p$.
$\punto$
\end{example}

\subsection{Pattern utility computation}
\label{sub:pattern-utility-computation}
Recall that a pattern is a (possibly ordered) set of items and it may occur in a certain number of transactions.
The generalization of utilities in facets, and the structuring of the database at different abstraction levels pave the way to more advanced and diversified computations of utilities. 

\paragraph{Intra-pattern utility.}
First of all, given a pattern $P$, composed of a set of $r$ items  and occurring in a transaction $T_p$, all the item utility vectors of items $i \in P$ must be merged into a unique item utility vector $IU$. In this process, internal utilities of items for $T_p$ can be taken into account. Formally, given the set $\text{\textit{IUS}}=\{IU_{1}, \ldots IU_{r}\}$ of item utility vectors associated with each item $i \in P$, let's define the \emph{intra-pattern} utility function $ip$ which takes as input the pattern $P$, the transaction $T_p$ and the associated set of item utility vectors $\text{\textit{IUS}}$, and generates a unique item utility vector for the pattern occurrence:
\[
IU_{T_p}= ip(P,T_p,\text{\textit{IUS}})
\]

\begin{example}\label{ex:intrapattern}
An example of \emph{intra-pattern} utility function is the following:
\[
IU_{T_p}=ip(P,T_p,\{IU_{1}, \ldots IU_{r}\})=
\]
\[
[\sum_{i \in [1..r]} (IU_{i}[1] \times q(i,T_p)), \sum_{i \in [1..r]} (IU_{i}[2] \times q(i,T_p)),    \ldots, \sum_{i \in [1..r]} (IU_{i}[l] \times q(i,T_p))]
\]

\noindent Depending on the context of interest, the combination of the utilities across the single facets can be carried out with functions different from the SUM. As an example, MAX, MIN, or AVG operators can be applied across the same facet of the different items in the pattern. Interestingly, if $r=1$ it corresponds to the classical definition. $\punto$
\end{example}

\paragraph{Pattern utility functions.}
Now, given a pattern $P$ occurring in a transaction $T_p$, the corresponding \emph{occurrence utility vector} $OccU_{T_p}$ is obtained by juxtaposing the item, transaction, object and container utility vectors:
\[
OccU_{T_p}=[IU_{T_p}, TU_{T_p}, OU_{T_p}, CU_{T_p}]=
\]
\[
[iu_1, \ldots, iu_l, tu_1, \ldots, tu_m, ou_1, \ldots, ou_n, cu_1, \ldots, cu_o]
\]
Here, for the sake of simplicity, we refer to $OU_{T_p}$ (resp., $CU_{T_p}$) as the object (resp., container) utility vector of the object (resp., container) containing transaction $T_p$. 

Given the set $T_P$ of transactions containing occurrences of the pattern $P$, the \emph{pattern utility vector} $U_P$ is obtained as the collection of all the occurrence utility vectors of $P$:
\[
U_P = \bigcup_{T_p \in T_P} OccU_{T_p}
\]

\begin{example}[\emph{(Running example)}]\label{example:3}
Let us continue the running example. Consider the pattern $P=\{\textsf{paper},\textsf{reproducibility}\}$, and the set of transactions $T_P = \{S_2,S_4\}$ in which it occurs. To proceed with the pattern utility computation, we first should compute its intra-pattern utility by merging the item utility vectors into a unique item utility vector. However, in the paper reviews use case, items have no facets. Hence, a simple intra-pattern utility function which returns an empty list can be exploited, that is $IU_{S_2} = IU_{S_4} = []$. The corresponding occurrence utility vector $OccU_{S_2}$ and $OccU_{S_4}$ can now be derived as the juxtaposition of unique transaction, object and container utility vectors:
\begin{gather*}
    OccU_{S_2} = [IU_{S_2}, TU_{S_2}, OU_{R_2}, CU_{C_1}] = [1,0,1,0,-1,-1,0,0,4,3,0] \\
    OccU_{S_4} = [IU_{S_4}, TU_{S_4}, OU_{R_3}, CU_{C_2}] = [0,1,-1,1,1,0,1,1,9,3,1]
\end{gather*}

\noindent 
Then, the pattern utility vector $U_P$ consists of the occurrence utility vectors $OccU_{S_2}$ and $OccU_{S_4}$, that is:

\begin{gather*}
U_P = \{OccU_{S_2}, OccU_{S_4}\} =
\{[1,0,1,0,-1,-1,0,0,4,3,0],[0,1,-1,1,1,0,1,1,9,3,1]\}
\end{gather*}
 $\punto$
\end{example}

Now, from $U_P$ we can virtually build a matrix where each row represents a utility vector associated with an occurrence of $P$ and each column represents a facet of $P$. The utility $u$ of the pattern $P$ can be then obtained as an arbitrary combination of the values in $U_P$, using a function $u(P)$.

In order to formalize function $u(P)$ we distinguish between formulas that operate \emph{by row} (we call them \emph{horizontal first} and we refer to them as $f_h$), formulas that operate \emph{by column} (we call them \emph{vertical first} and we refer them as $f_v$), and formulas that operate on the whole data at once (we call them \emph{mixed} and we refer them as $f$). 

Formally, utility of $P$ can be classified in:

\begin{itemize}
    \item \emph{Horizontal first}; it first combines utilities of the various facets in each occurrence (by row) and then combines the values across all occurrences (by column): 
    $u(P)=f_v(f_h(U_P))$
    
    \item \emph{Vertical first}; it first evaluates utilities on a facet basis across the  occurrences (by column) and then combines the obtained values across the facets (by row): $u(P)=f_h(f_v(U_P))$
    
    \item \emph{Mixed}; it combines the values in $U_P$ at once: $u(P)=f(U_P)$

\end{itemize}

\noindent Observe that $u(P)$ is a single number, whereas intermediate computations may provide sets of values. 

Both \emph{Horizontal first}, \emph{Vertical first}, and \emph{Mixed} utility functions may be further classified in: 

\begin{itemize}

    \item \emph{inter-transaction} utility: functions that combine item and transaction utilities; 
    
    \item \emph{pattern-vs-object} utility: functions that compute the utility of the pattern by correlating one or more item or transaction utility facets with one or more object utility facets;
    
     \item \emph{pattern-vs-container} utility: functions that correlate one or more item or transaction utility facets with one or more container utility facets.
\end{itemize}

The list of potentially interesting functions is virtually infinite. A non-exhaustive set of functions for some of the classes introduced above is provided next. Their specific semantics is strongly dependent on the application context.
\begin{itemize}
 \item \emph{Horizontal first} - \emph{inter-transaction}: 
     \begin{itemize}
         \item Filter \& Sum: Filters one single facet first and then sums the obtained values\footnote{Observe that this is equivalent to the classical HUPM scenario.}.
         \item Filter \& Times: Filters one single facet first and then multiplies the obtained values\footnote{Observe that,  if the facet represents a probability, this can be used to compute a combined probability.}.
         \item Coherence degree: returns a predefined value if (a subset of) facets show coherent values and then computes the percentage of pattern occurrences containing this value; an index of coherence can be the fact that all facets are positive or all facets are negative.
     \end{itemize}
     \item \emph{Horizontal first} - \emph{pattern-vs-object} / \emph{pattern-vs-container}:
     \begin{itemize}
         \item Disagreement degree: returns a predefined value if the value of an item/transaction facet disagrees with that of an object/container facet and then computes the percentage of pattern occurrences containing this value.
     \end{itemize}
  
\item \emph{Vertical first} - \emph{inter-transaction}: 
     \begin{itemize}
         \item Max \& Sum: computes the maximum for each facet across transactions first and then sums the obtained values. To understand the applicability of this function, consider a context in which each facet is a rating of a certain aspect of a transaction (e.g. shipping speed, reliability,  etc.); this function makes it possible to find the patterns with the highest overall ratings on all facets.
         \item Std \& Max: computes the standard deviation of each facet across transactions first and then takes the maximum value among facets. 
     \end{itemize}
     \item \emph{Vertical first} - \emph{pattern-vs-object}:
     \begin{itemize}
         \item Mixed Coherence degree: for each facet, computes the fraction of transactions with positive (resp., negative) values, then filters on an item/transaction facet and on an object facet and multiplies the corresponding fractions. 
     \end{itemize}

\item \emph{Mixed} \emph{pattern-vs-object} / \emph{pattern-vs-container}: 
     \begin{itemize}
         \item Pearson Correlation~\citep{Pearson*1895}: computes the correlation between one of the item/transaction facets and one of the object/container facets. It can be used to measure the agreement/disagreement about information at pattern/transaction level and object/container level. 
     \item Multiple Correlation~\citep{Lebe*03}: computes the correlation between two or more item/transaction facets and one of the object/container facets. It can be used to measure how well a given object/container facet can be predicted using a linear function of a set of item/transaction facets.
  \end{itemize}

\end{itemize}

\begin{example}[\emph{(Running example)}]
Let us consider the pattern utility vector $U_P$ for the pattern $P=\{\textsf{paper},\textsf{reproducibility}\}$ shown in  Example~\ref{example:3}. We now give examples for some of the classes of utility functions introduced above.

We start by giving a simple example for Horizontal and Vertical first functions. We consider an inter-transaction function involving Filter \& Max. More in detail, let $f_h = \text{filter}(\cdot)$ and $f_v = \text{max}(\cdot)$ be the Filter and Max functions, respectively. Then, we have $u(P) = f_v(f_h(U_P)))$. Assume we want to focus on the rating facet of each object utility vector, therefore $u(P)$ consists in filtering the rating facet (by means of $f_h$) and then selecting the maximum value of such facet (by means of $f_v$). In detail, $f_h(U_P) = \text{filter}(U_P) = \{[4], [9]\}$, where $[4]$ (resp., $[9]$) indicates the rating facet of $OccU_{S_2}$ (resp., $OccU_{S_4}$) represented as a singleton. Then, by applying $f_v$ across each occurrence, we obtain $u(P) = f_v(f_h(U_P)) = \text{max}(\text{filter}(U_P)) = \text{max}(\{[4],[9]\}) = 9$. In this case, the same result is obtained by applying these functions in a vertical first fashion, e.g., $u(P) = f_h(f_v(U_P))$.

Instead, if we consider a mixed pattern-vs-object utility function, an interesting example is the computation of the Pearson correlation coefficient. Let us define $f = r_{xy}$, where $r_{xy}$ is the classic Pearson correlation coefficient. Also, suppose that $x$ indicates the Clarity aspect value present in the sentence and $y$ indicates the rating facet of the review, across all occurrence utility vectors in $U_P$. 
According to $U_P$, we have $x = [0,1]$ and $y = [4,9]$, and by applying the pattern-vs-object utility function $f$ we obtain $f(U_P) = r_{xy} = 1.0$, indicating a perfect positive correlation between Clarity aspect and the rating facets for the pattern $P=\{\textsf{paper},\textsf{reproducibility}\}$.
$\punto$
\end{example}

\subsection{Pattern masks.}
As a final extension, we introduce \emph{pattern masks}, i.e., properties that patterns (resp., pattern occurrences) must satisfy in order to be considered valid patterns (resp., occurrences). 
Simple examples of pattern masks are given below. Clearly, other kinds of masks can be defined, depending on the context of interest. 

\begin{itemize}
    \item  Mask on pattern size: it constrains the number of items in a pattern into a given range. 

    \item Mask on pattern (external) properties: it constrains items in the pattern to satisfy certain properties. 
    As an example, if items are words from review sentences, each word can be associated with a Part of Speech (POS) tag, and it can be interesting to constrain each pattern to contain at least a \emph{noun}, a \emph{verb}, and an \emph{adjective}.   

\end{itemize}

\subsection{Definition of extended high utility patterns.} 
\label{sec:problemdefinition}
In the classical problem, a high utility pattern is detected disregarding its support on the database focusing on its utility only. In the new setting we are proposing in this work, we must consider that there are utility functions which may have low significance in presence of few transactions supporting the pattern. 
As an example, in order to properly compute the Pearson correlation at least two, but preferably more, data points are necessary.
As a consequence, in order to provide a framework as general as possible, our problem formulation includes the definition of a minimum support for the pattern, in order to detect high utility patterns; obviously, this threshold can be set to 1,  meaning that a pattern should appear at least once, whenever a minimum support is not relevant.

\paragraph{Problem definition.} Given a pattern $P$ containing a set of items, and a pattern mask $M$, $P$ is an extended high utility pattern if $P$ and its occurrences satisfy $M$, its utility $u(P)$ is greater than a utility threshold $th_u$, and it occurs in at least $th_f$ transactions. The problem of extended high utility pattern mining is to discover all extended high utility patterns in a database $D$. 

\section{Design of the ASP Approach}
\label{sec:asp}
As previously pointed out, one of the main objectives of this work is to provide as much flexibility as possible in the definition of what is a valid pattern. In what follows, we  provide an encoding as general as possible for the framework introduced in Section \ref{sec:ehupm-framework}; this can be specialized for specific settings and, indeed, we also show how it can be specialized for the paper reviews use case introduced as running example in Section \ref{sec:ehupm-framework}. 

It is important to point out again that coding complex formulae for pattern utility, such as some of those described in Section \ref{sec:ehupm}, may in general not be easy (or even  impossible) to achieve with a purely declarative approach; think, for instance, about the computation of the standard deviation or the multiple correlation coefficient. Similar difficulties may arise when it is necessary to handle real or negative numbers.
To overcome this issue, we resort to recent extensions of ASP systems, such as DLVHEX~\citep{Eiter*16,Eiter*18}, clingo~\citep{Gebser*19}, and WASP \citep{DoRi20}, which allow the integration of external computation sources, usually written in Python, within the ASP program. 
%
The ASP standardization group has not yet released standard language directives for such features; in order to present our ASP formalization in this section, we refer to syntax and semantics of DLVHEX~\citep{Eiter*16}, which is more intuitive. In Section \ref{sec:application} we also show a complete implementation in WASP \citep{DoRi20} that exploits constraint propagators.

The general encoding is presented in Listing~\ref{alg:asp-general}; it is structured in separate parts, in such a way that the various aspects of the problem can be modelled, and changed, with local  modifications. 

The first part (lines $1$-$9$) recalls the expected schema for input facts to simplify the reading of the code. Observe that attributes {\tt Position} and {\tt Q} for predicate {\tt item} will be needed only if item position within the transaction and its internal utility are actually needed for the problem at hand (see Section \ref{sub:facets-utilities}). Similarly, the number of facets for item, transaction, object and container utility vectors will be specific to the application scenario; the corresponding predicates may be omitted if no facets are available for them.  
Thresholds for pattern frequency ($th_f$) and utility ($th_u$) must be provided as input as well with facts {\tt occurrencesThreshold} and {\tt utilityThreshold} (line $11$).

Rule prototype in line $13$ allows items to be pre-filtered, according to users' background knowledge. Think for instance to items with a price lower than a certain threshold which are certainly not relevant for the analysis. Such filters can be very application-specific and we therefore leave this part open to user specification. Predicate {\tt usefulItem} is true only for unfiltered items. 

The second part of the formulation considers the generation of candidate patterns and the verification of their minimum frequency.  
The chosen encoding generates one answer set for each pattern; this simplifies both pattern representation and the application of user-defined constraints to candidate patterns.
As pointed out in Section \ref{sec:problemdefinition}, in the classical HUPM context minimum support is not considered at all as a constraint on pattern validity (only utility is actually considered); in this setting, all combinations of items would be candidate patterns. However, we decided to keep the frequency constraint in order to cope with specific application scenarios. 
To identify the set of candidate patterns with a minimum support, we build upon the effective and elegant solution already presented in \cite{Jarvisalo11}, which has been adapted to our context (lines $15$-$18$).  Here, the predicate {\tt inCandidatePattern} is true for an item $i$ only if $i$ is a useful item and belongs to a frequent pattern. The predicate {\tt inSupport} is true for a transaction $t$ if it supports all the items in the candidate pattern. The lack of support for some items is determined by the {\tt conflictAt} predicate. We also use the support predicate {\tt contains} to project the first two variables of {\tt item}. Observe that minimum frequency is checked directly in the choice rule (line 15). 

The third part of the formulation (lines $20$-$21$) generates the unique item utility vector for the pattern occurrence; in particular, predicate {\tt patternItemUtilityVectors} is true for each item $i$ in the candidate pattern supported by a transaction $t$, and collects the corresponding facets and internal utility. Predicate {\tt intraPatternUtilityVector} then instantiates the unique item utility vector for the corresponding transaction by the application of the external function {\tt computeIntraPatternUtility}. Observe that, generally, this computation involves simple sum-product operations; in this case, it might be more convenient to express the  formula directly within the rule. Moreover, we point out that, when items have no facets and internal utility, this part of the encoding can be omitted.   

The fourth part of the formulation (lines $23$-$25$) builds the occurrence utility vector (line 23) and checks utility criteria on candidate patterns (line 25). Here, the different database layers introduced by the framework are taken into account and the full power of external functions is exploited. Observe that when using WASP constraint propagators, the constraint expressed in line $25$ is replaced by a propagator.

Finally, the last part of the formulation (lines $27$-$29$) is devoted to express pattern masks; this part is very application specific and we show a mask on pattern size just as a simple example. Here, users' background knowledge may play a relevant part. As an example, if items are words and we know for each word its part-of-speech tag, a more elaborated mask can be defined requiring that a pattern must contain at least a noun, a verb, and an adjective.

\lstdefinelanguage{asplang}{
    morecomment=[l]{\%}
}
\lstset{language=asplang, basicstyle=\scriptsize\ttfamily, breaklines=true, numbers=left, numberstyle=\footnotesize, columns=fullflexible, morecomment=[l][\color{blue}]{\%}}
\begin{lstlisting}[frame=single, float=ht!, caption={A general ASP encoding for the e-HUPM problem}, label={alg:asp-general}]
%%%   Input schema:
% container(ContainerId)
% object(ObjectId, ContainerId)
% transaction(Tid, ObjectId)
% item(Item, Tid, Position, Q)
% itemUtilityVector(Item, I1, ..., Il)
% transactionUtilityVector(Tid, T1, ..., Tm)
% objectUtilityVector(ObjectId, O1, ..., On)
% containerUtilityVector(ContainerId, C1, ..., Co)
%%%   Parameters
occurrencesThreshold(...). utilityTreshold(...).
%%%   Item pre-filtering
usefulItem(I):- item(I,_,_,_),....any condition on the items.
%%%   Candidate pattern generation and check on minimum support
{inCandidatePattern(I)}:- usefulItem(I), #count{Tid : inSupport(Tid), contains(I,Tid)}=N, N >= Tho, occurrencesThreshold(Tho).
inSupport(Tid):-  transaction(Tid,_), #count{I : usefulItem(I), conflictAt(Tid,I)}=0.
conflictAt(Tid,I):- inCandidatePattern(I), transaction(Tid,_),  not contains(I,Tid).
contains(I,Tid):- item(I,Tid,_,_). 
%%%   Intra-pattern Utility computation (needed only if items have utilities)
patternItemUtilityVectors(Tid,Item,I1,...,Il,Q):- inCandidatePattern(Item), itemUtilityVector(Item,I1,...,Il), inSupport(Tid), item(Item,Tid,Position,Q).
intraPatternUtilityVector(Tid,I1,...,Il):- &computeIntraPatternUtility[patternItemUtilityVectors](Tid,I1,...,Il).
%%%   Pattern Utility computation 
occurrenceUtilityVector(Tid,I1,...,Il,T1,...Tm,O1,...On,C1,...,Co):- inSupport(Tid), intraPatternUtilityVector(Tid,I1,...,Il), transactionUtilityVector(Tid,T1,...,Tm),  transaction(Tid, ObjectId), objectUtilityVector(ObjectId,O1,...,On), object(ObjectId, ContainerId), containerUtilityVector(ContainerId,C1,...,Co).
%%%   The following constraint is implemented as a constraint propagator in WASP
:- &computeUtility[occurrenceUtilityVector](U), U < Thu, utilityTreshold(Thu).
%%%   Pattern mask (e.g.) on size
minLength(...). maxLength(...). 
:- #count{T : inCandidatePattern(T)} < L, minLength(L). 
:- #count{T : inCandidatePattern(T)} >  L, maxLength(L).
\end{lstlisting}

\begin{figure}
    \centering
        \includegraphics[width=1.0\linewidth]{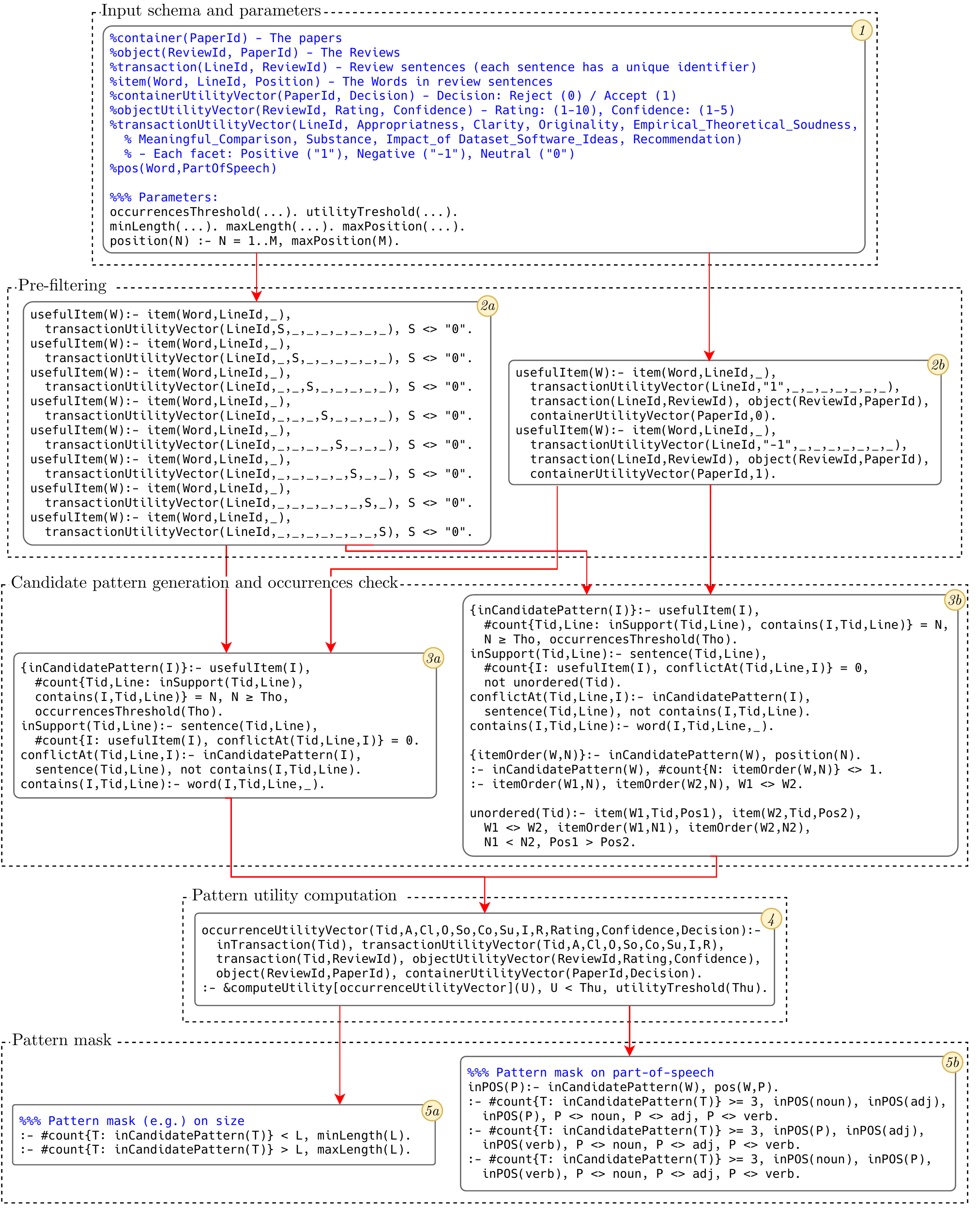}
        \caption{Modular composition of ASP subprograms for the paper reviews use case introduced as running example}
    \label{fig:tplp-ehupm-fig}
\end{figure}

The general ASP encoding presented in Listing \ref{alg:asp-general} can be specialized in different ways depending on the application context and on the focus of the analysis. In what follows, we show a complete formulation of the proposed approach for the  paper reviews use case introduced as running example in Section \ref{sec:ehupm-framework}; in particular, we show how the various parts of the general program described above, which we call {\em Blocks} in the following to simplify the presentation, can be modelled by different ASP code portions that can be combined in many different ways. In our opinion, this makes evident the flexibility provided by our hybrid solution combining ASP and Python, where the main ability of declarative programming, which allows programmers to define the what and not the how, is fully exploited in the definition and combination of the various Blocks, whereas the potential of Python to switch between different complex utility functions is exploited by external functions.  

Figure \ref{fig:tplp-ehupm-fig} shows the proposed formulations, where the diverse Blocks are numbered in order to simplify the presentation. In particular, Block 1 simply defines the schema of the input dataset, along with expected input parameters. Observe that, for the paper reviews use case, no utility is associated with items; consequently, there is no {\tt itemUtilityVector} and, thus, no intra-pattern utility computation is needed. For this reason, the proposed formulation does not include a Block for intra-pattern utility computation.

As far as the pre-filtering part is concerned, Block 2a and Block 2b show two different alternatives. In the code of Block 2a, only words associated with at least one positive or negative sentiment value for the sentences they appear in are considered useful for the analysis; Block 2b shows a pre-filtering that includes only words for which a disagreement between Appropriateness and Decision exists. Obviously many other pre-filtering rules can be defined for the problem at hand.  

Block 3a and Block 3b refer to the candidate pattern generation and occurrences check part. In particular,  Block 3a reproduces the code already described in the general encoding, whereas Block 3b shows how easy it can be  with ASP to switch from the classical scenario to a very different problem, in this case sequence pattern mining. In particular, in this formulation, the classical code is slightly modified in such a way that the order of items within sentences becomes relevant, and an occurrence of an item in a sentence is validated only if all items occur precisely in the same order. It is interesting to observe that the switch of this Block alone makes it possible to tackle very different pattern mining problems, and that switching from a pattern mining problem to another one with an imperative programming solution would require to rewrite long code portions. 

Pattern utility computation, coded in Block 4, is standard; in this case the different potential utility functions are coded in Python by the external function {\tt computeUtility}.

Finally, Block 5a and Block 5b express two different ways of defining pattern masks. Block 5a defines a simple mask on pattern size, whereas Block 5b implements a more elaborated mask where,  if we assume to know  the part-of-speech tag of each word, it states that if a pattern contains at least three words, these must be at least a noun, a verb, and an adjective. Of course, again, one can imagine many other types of  masks.   

As a final consideration, it is noteworthy that even with a limited number of alternatives for each block, a wide variety of final encodings can be generated. As an example, the combination of Blocks 1-2a-3a-4-5a represents a full encoding; other possibilities are Blocks 1-2b-3a-4-5b or Blocks 1-2a-3b-4-5a, and many more. Given a pool of alternative encodings for each Block, even those unfamiliar with ASP could put up their own variants to provide maximum flexibility in data analysis.

\section{Application to ICU admission prediction for COVID-19 patients}
\label{sec:application}
In this section we present a possible application of the proposed approach in a biomedical context; in particular, we show how high utility patterns derived from a structured database can be exploited to compute correlations, and possibly make predictions, for ICU admission of confirmed COVID-19 cases, based on patients' blood results and vital signs. It is worth pointing out that the main objective of this section is not to provide a final solution to the problem, but to show with a proof-of-concept that our e-HUPM framework can actually help in relevant practical problems.

To this purpose, we exploit the COVID-19 dataset publicly available at \url{https://www.kaggle.com/S%C3%ADrio-Libanes/covid19}. 
The dataset contains almost 2000 anonymized data for confirmed COVID-19 patients; each patient in the database underwent several encounters. The dataset contains the following information: {\em (i)} patient demographic information, {\em (ii)} patient previous grouped diseases, {\em (iii)} if the patient was admitted to the ICU during the observation period (ICU=1) or not (ICU=0), and, for each encounter, {\em (iv)} 36 parameters for blood results, and {\em (v)} 6 parameters for vital signs. 
With respect to our e-HUPM model, a patient is an object, having a single facet, namely ICU; each visit is a transaction whose items are categorical data from {\em (i)} and {\em (ii)}  and whose 42 facets are numerical  data from  {\em (iv)} and {\em (v)}. No facets and internal utility are defined for items; there is no container layer for the dataset.

\lstdefinelanguage{asplang}{
    morecomment=[l]{\%}
}
\lstset{language=asplang, basicstyle=\scriptsize\ttfamily, breaklines=true, numbers=left, numberstyle=\footnotesize, columns=fullflexible, morecomment=[l][\color{blue}]{\%}}
\begin{lstlisting}[frame=single, float=ht!, caption={A full example of the ASP encoding for the ICU admission prediction for COVID-19 patients}, label={alg:asp-covid}]
%%% Input schema:
% object(PatientId)
% objectUtilityVector(PatientId, ICU)
% transaction(VisitId, PatientId)
% transactionUtilityVector(VisitId, ALBUMIN_MEAN, BE_ARTERIAL_MEAN, BE_VENOUS_MEAN, BIC_ARTERIAL_MEAN, BIC_VENOUS_MEAN, BILLIRUBIN_MEAN, BLAST_MEAN, CALCIUM_MEAN, CREATININ_MEAN, FFA_MEAN, GGT_MEAN, GLUCOSE_MEAN, HEMATOCRITE_MEAN, HEMOGLOBIN_MEAN, INR_MEAN, LACTATE_MEAN, LEUKOCYTES_MEAN, LINFOCITOS_MEAN, NEUTROPHILES_MEAN, P02_ARTERIAL_MEAN, P02_VENOUS_MEAN, PC02_ARTERIAL_MEAN, PC02_VENOUS_MEAN, PCR_MEAN, PH_ARTERIAL_MEAN, PH_VENOUS_MEAN, PLATELETS_MEAN, POTASSIUM_MEAN, SAT02_ARTERIAL_MEAN, SAT02_VENOUS_MEAN, SODIUM_MEAN, TGO_MEAN, TGP_MEAN, TTPA_MEAN, UREA_MEAN, DIMER_MEAN, BLOODPRESSURE_DIASTOLIC_MEAN, BLOODPRESSURE_SISTOLIC_MEAN, HEART_RATE_MEAN, RESPIRATORY_RATE_MEAN, TEMPERATURE_MEAN, OXYGEN_SATURATION_MEAN)
% item(VisitId, Value)
%%% Parameters
occurrencesThreshold(...). utilityThreshold(...). maxCardItemset(...).
%%% Pre-filtering: all items are useful
usefulItem(I):- item(_,I).
%%% We compute the Pearson correlation between each target facet and ICU. Each run differs only for 
%%% the projected facet and will detect patterns with a high correlation between that facet and ICU.
transactionUtilityVectorP(Tid,T):- transactionUtilityVector(Tid, _, _, _, _, _, T, _, _, _, _, _, _, _, _, _, _, _, _, _, _, _, _, _, _, _, _, _, _, _, _, _, _, _, _, _, _, _, _, _, _, _, _).
%% Candidate pattern generation and check on minimum support
{inCandidatePattern(I)}:- usefulItem(I), #count{Tid: inSupport(Tid), contains(I,Tid)}=N, N>=Tho, occurrencesThreshold(Tho).
inSupport(Tid):- transaction(Tid,_), #count{I: usefulItem(I), conflictAt(Tid,I)}=0.
conflictAt(Tid,I):- inCandidatePattern(I), transaction(Tid,_), not item(Tid,I).
%% The utility of the whole occurrence is the juxstaposition of all the required values
occurrenceUtilityVector(Tid,ICU,Target):- inSupport(Tid), transaction(Tid,PatientId),
    objectUtilityVector(PatientId,ICU), transactionUtilityVectorP(Tid,Target).
%% The following is implemented as a constraint propagator in WASP and must be commented here
%:- &computeUtility[occurrenceUtilityVector](U), U < Thu, utilityThreshold(Thu).
%% size of each pattern is at most M
cardItemset(N):- #count{I : inCandidatePattern(I)} = N.
:- cardItemset(N), maxCardItemset(M), N > M.
:- cardItemset(N), N < 1. 
\end{lstlisting}

\lstdefinelanguage{python}{
    morecomment=[l]{\#}
}
\lstset{language=python, basicstyle=\scriptsize\ttfamily, breaklines=true, numbers=left, numberstyle=\footnotesize, columns=fullflexible, morecomment=[l][\color{blue}]{\#}}
\begin{lstlisting}[frame=single, float=ht!, caption={The core Python function of the constraint propagator exploited in WASP}, label={alg:py-covid}]
def compute(answer_set):
    global aspvars, threshold, icu_values, target_values
    icu_values, target_values = [], []
    for x in answer_set:
        if x < 0: continue
        if x not in aspvars: continue
        
        (_, icu, target) = aspvars[x]
        icu_values.append(icu)
        target_values.append(target)

    # Compute Pearson via scipy library
    pearson_value = stats.pearsonr(icu_values, target_values)[0]
    
    # Check if computed value is valid
    sat = abs(pearson_value) >= threshold
    return (pearson_value, not sat)
\end{lstlisting}

To give a full example of application of our framework, Listing~\ref{alg:asp-covid} reports the ASP program specialized to this context, where WASP propagators are used to implement the constraint in line $22$. The Listing~\ref{alg:py-covid} shows an excerpt of the propagator implementation; here, the most important thing to note is the simplicity with which the Pearson correlation can be calculated (see line $13$ of Listing~\ref{alg:py-covid}). It is beyond the scope of this paper to explain all the details of the implementation of propagators. The interested reader is referred to \citep{DoRi20} for the literature on propagators and to  \url{https://www.mat.unical.it/~cauteruccio/tplp-ehupm/} for the complete implementation of propagators for this problem.

The first objective of this analysis was to find patterns, derived from patient demographic information and patient disease groups, showing a significant Pearson correlation between each of the 42 facets and ICU outcome.  
As an example, for the pattern: (Gender:Male, AgePercentile:60, DiseaseGroup5:YES) we found a Pearson correlation of -0.75 between Oxygen Saturation values and ICU; similarly, for the pattern (Gender:Female, Immunocompromised:NO, DiseaseGroup5:YES, DiseaseGroup6:NO, Hypertension:YES) we found a Pearson correlation of -0.82 between Hemoglobin values and ICU. 
More generally, we found 590 (resp., 1422) patterns showing an absolute value of the Pearson correlation greater than 0.5 between Oxygen Saturation (resp., Hemoglobin) values and ICU occurring in at least 10 transactions in the dataset. Similar number of patterns have been derived also for the other facets. 
Note that, to switch from one facet to another, it was sufficient to change the attribute to be projected in line $13$ of Listing~\ref{alg:asp-covid}.
%

\begin{figure}
    \centering
    \begin{subfigure}{.5\textwidth}
        \centering
        \includegraphics[width=0.85\linewidth]{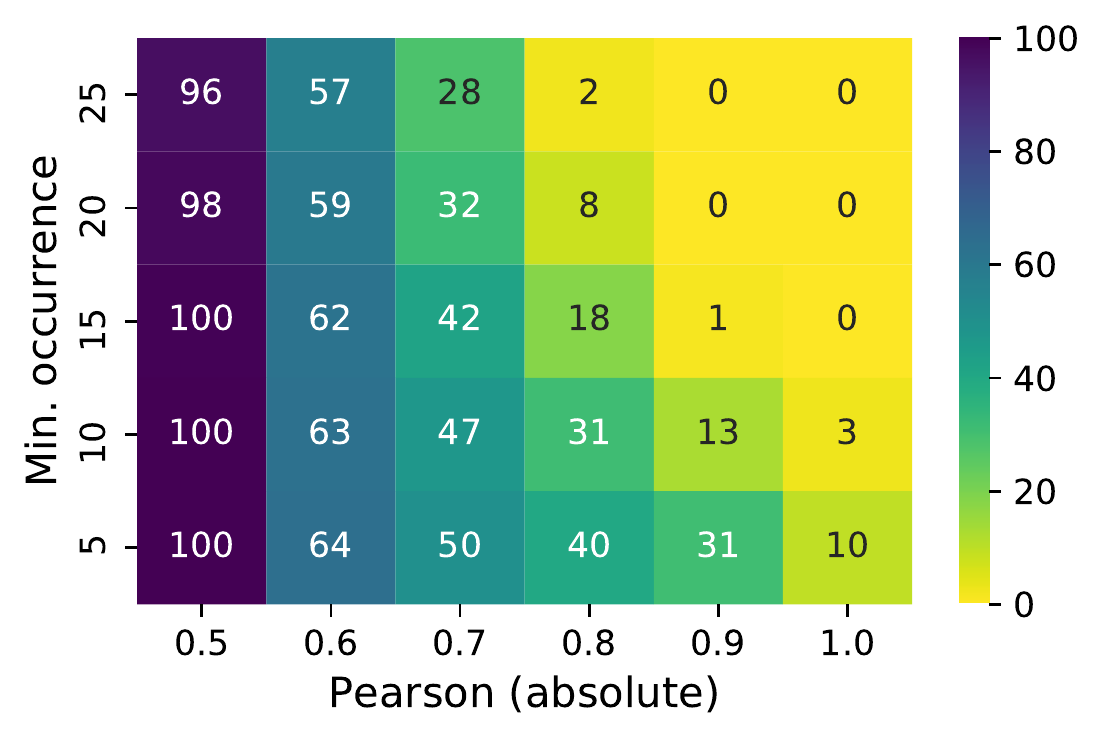}
        \caption{}
        \label{fig:occurrence_vs_pearson}
    \end{subfigure}%
    \begin{subfigure}{.5\textwidth}
        \centering
        \includegraphics[width=0.85\linewidth]{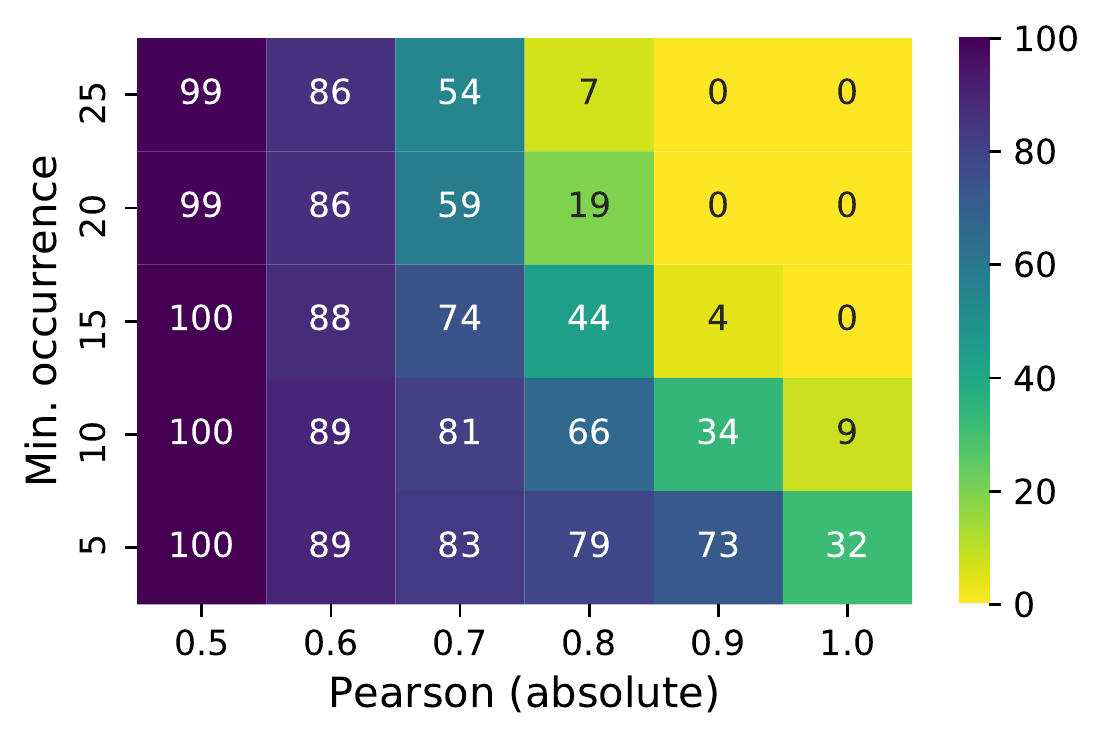}
        \caption{}
        \label{fig:combinations_occurrence_vs_pearson}
    \end{subfigure}
    \caption{Percentage of transactions (a) and Percentage of combinations of patient attributes (b) covered by valid patterns}
    \label{fig:occ_vs_pearson}
\end{figure}

As a preliminary analysis of obtained results, in light of exploiting them for ICU prediction, we analyzed the fraction of transactions in the database supported by at least one valid pattern, and the percentage of combinations of patients attributes covered by at least one valid pattern. This last has been measured as the fraction of combinations that can be extracted from available data which are also supported by at least one valid pattern. 
Observe that, having a high fraction of covered transactions (resp., combinations) might be important in order to improve the chances to have a personalized predictive model for each upcoming unknown patient. However, setting very low minimum thresholds just to increase the number of valid patterns may, in principle, be counterproductive for their subsequent use in prediction tasks. It is then important to study these properties for a better understanding of the next steps of analysis. Results obtained from this analysis are shown in Figure \ref{fig:occ_vs_pearson} for several values of minimum occurrence and minimum correlation thresholds. From the analysis of this figure, we can observe that even a slight increase of the minimum correlation has a strong impact on the fraction of transactions/combinations covered by valid patterns; an increase of the minimum occurrence, instead, has a lower impact. Interestingly, too stringent thresholds end up to completely miss all transactions/combinations in the dataset and, clearly, these should be avoided. As pointed out in \citep{ScBoSc18} a Pearson between 0.40 and 0.69 represents moderate correlations and, as such, a correlation of 0.5 should not be neglected. Similarly, correlations above 0.7 represent strong or very strong relationships; these might be considered very reliable but, as observed from this experiment, exploiting these values may end up in discarding too much data.    

At the end of the first phase, we have a set ${\cal P}=\{\langle p_i , \pi_j \rangle \}$  of patterns $p_i$ for each facet $\pi_j$ corresponding to a target parameter (as an example, Oxygen saturation, Hemoglobin, etc.). Now, each $\langle p_i , \pi_j \rangle $ is characterized by a number $\nu_{ij}$ of transactions supporting it and a Pearson correlation $\gamma_{ij}$. It is then possible to derive a linear regression model $\mu_{ij}(\cdot)$ between $\pi_j$ and ICU for $p_i$. Given the value $v_{jk}$ for the facet $\pi_j$ of a generic (possibly unseen) transaction $t_k$ supported by  $\langle p_i , \pi_j \rangle$, we can then apply $\mu_{ij}(v_{jk})$  to estimate the value of ICU. $\mu_{ij}(v_{jk})$ is a value in the real interval $[0,1]$ and constitutes a building block for the overall prediction process which is described next.  

Given a generic unseen transaction $t_k$, let first identify the set ${\cal P}_{t_k} \subseteq {\cal P}$ of patterns in ${\cal P}$ contained also in $t_k$. Observe that, in principle, ${\cal P}_{t_k}$ might be void; in this case, no prediction can be carried out. For each  $\langle p_i , \pi_j \rangle \in {\cal P}_{t_k}$  and the corresponding value $v_{jk}$ in $t_k$ let apply the regression model $\mu_{ij}(v_{jk})$ previously constructed, in order to get an estimation of ICU for the patient corresponding to $t_k$ based on $v_{jk}$. Now, since there can be more patterns in ${\cal P}_{t_k}$ for $t_k$, it is possible to produce a more accurate estimation by combining the predictions obtained from all the patterns.

There are several ways in which we can combine the various predictions. Since each $\langle p_i , \pi_j \rangle$ is characterized by both a Pearson correlation $\gamma_{ij}$ and a number of transactions $\nu_{ij}$ supporting it, we can use both these values as weights in an average mean of all obtained predictions. In fact, a very high correlation indicates a very good predictive capability; however, if this correlation is not supported by many transactions, it might be unreliable for unseen transactions. On the contrary, a high number of transactions supporting the pattern indicates a good support for the prediction, but obtaining a high correlation from several transactions is, in general, more complicated. By combining these two factors we can weigh each prediction in a more accurate way. In particular, the formula exploited to estimate ICU for the patient corresponding to the transaction $t_k$ is:
\[
ICU_k=  \frac{\sum_{ij} \mu_{ij}(v_{jk})\cdot\gamma_{ij}\cdot\nu_{ij}}{\sum_{ij}\gamma_{ij}\cdot\nu_{ij}}
\]
Observe that $ICU_k$ is a value in the real interval $[0,1]$; in order to finalize the estimation, we carry out a rounding up of its value.

\begin{table}
   \centering
    \caption{Accuracy (left) and missing rate (right)}
    \label{tab:prec-missing}
   {\small
    \begin{tabular}{ccccccc}
        \topline
        \textit{Min. Occ.} & \multicolumn{6}{c}{\textit{Pearson (absolute)}}
        \midline
         & 0.5 & 0.6 & 0.7 & 0.8 & 0.9 & 1.0 
        \midline
        5 & 0.73 & 0.75 & 0.76 & 0.74 & 0.78 & 0.77 \\
        10 & 0.71 & 0.73 & 0.73 & 0.72 & 0.79 & 0.78 \\
        15 & 0.72 & 0.71 & 0.7 & 0.71 & 0.83 & - \\
        20 & 0.71 & 0.71 & 0.71 & 0.67 & - & - \\
        25 & 0.71 & 0.7 & 0.71 & 0.7 & - & -
        \botline
    \end{tabular}
    \begin{tabular}{ccccccc}
        \topline
        \textit{Min. Occ.} & \multicolumn{6}{c}{\textit{Pearson (absolute)}}
        \midline
         & 0.5 & 0.6 & 0.7 & 0.8 & 0.9 & 1.0 
        \midline
        5 & 0.15 & 0.39 & 0.48 & 0.55 & 0.61 & 0.74 \\
        10 & 0.15 & 0.39 & 0.5 & 0.6 & 0.72 & 0.78 \\
        15 & 0.16 & 0.4 & 0.53 & 0.68 & 0.79 & 1.0 \\
        20 & 0.17 & 0.41 & 0.6 & 0.75 & 1.0 & 1.0 \\
        25 & 0.18 & 0.43 & 0.62 & 0.79 & 1.0 & 1.0
        \botline
    \end{tabular}}
\end{table}

In order to test the validity of the approach we carried out a 5-fold cross-validation where, in each run, 80\% of tuples are used for extracting patterns and building the prediction model, and 20\% of tuples to compute predictions.  A prediction is considered exact if its value corresponds to the ICU value present in the database for the corresponding patient. This allows us to measure the Accuracy of the approach as the fraction of exact results vs the total ones. Table~\ref{tab:prec-missing} (left) shows the average accuracy obtained among all the runs varying the minimum allowed frequency and the Pearson correlation. Since a transaction might not be supported by valid patterns, Table~\ref{tab:prec-missing} (right) shows also the missing rate, i.e., the percentage of transactions with no predictions.  We also computed the variance for the obtained results on Accuracy, which ranges from a minimum of 0.0015 to a maximum of 0.057.

From the analysis of this table we can observe that the Accuracy is quite stable and above 70\% across the various combinations of parameters, with a slight increase as the Pearson increases. However, looking at the missing rate, we can observe that increasing values of the correlation significantly increases also the missing rate. If we observe Figure \ref{fig:occ_vs_pearson} there are actually threshold values for which the missing rate is very high (even 1, when no predictions are available); the Accuracy obtained for these configurations, although higher than that of the others, is of less significance since these configurations only allow us to obtain predictions on a small set of data.
Reasonable values for the missing rate are obtained for a minimum absolute Pearson equal to 0.5, which corresponds also to a satisfying Accuracy. The quite stable behaviour of the Accuracy can be motivated by the weighed formula we adopted for computing the predictions, which is able to adapt either to a high number of supported patterns and transactions or to a high Pearson with less support. Summarizing, it turns out that it is better to have a good coverage with a moderate correlation than a very strong correlation with a low coverage. 

\section{Experimental Evaluation}
\label{sec:experiments}

In this section we show and discuss the results of several experiments we carried out in order to assess the effectiveness of the proposed approach. We focus on the paper reviews use case introduced as running example and we consider a {\em quantitative} analysis, aiming to characterize the applicability of the approach in terms of performances, and a {\em qualitative} analysis, aiming to assess the quality of results.  

The full dataset exploited in these experiments includes 814 papers, 1148 reviews and 2230 annotated sentences, containing overall 15124 distinct words.  The complete dataset, expressed in ASP format, is available  at \url{https://www.mat.unical.it/~cauteruccio/tplp-ehupm/}.
Other properties are shown in Table \ref{tab:papers-vs-hupm}.
All the experiments have been performed on a 2.3GHz MacBook computer (Intel Core i9) with 16 GB of memory. The data preprocessing pipeline was implemented in Python 3.9 and exploited the spaCy (\url{https://spacy.io/}) library. In order to run our ASP programs, we used clingo version 5.4.0 with the flag \texttt{--mode=gringo} as the grounder and WASP version 2.0 compiled via clang (version 11.0.3) with Python 3.9 support as the solver.

\subsection{Quantitative Analysis}
\label{sub:quantitative}
\begin{figure}
    \centering
    \begin{subfigure}{.5\textwidth}
        \centering
        \includegraphics[width=0.9\linewidth]{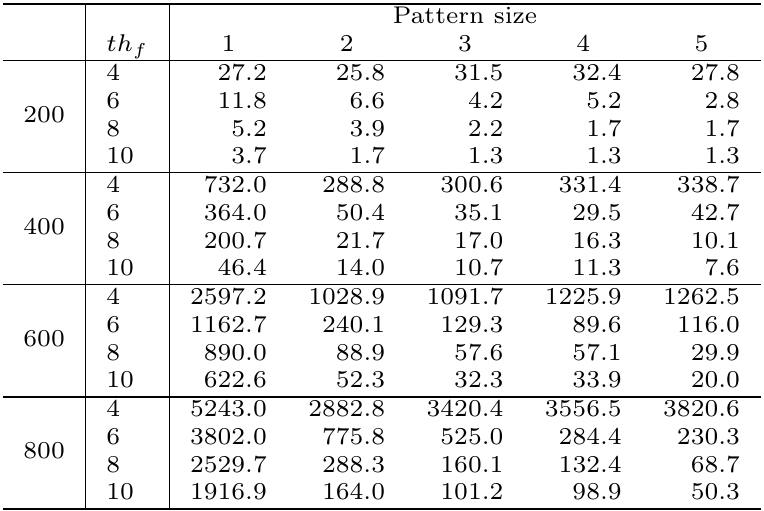}
        \caption{}
        \label{fig:npapers_vs_avgtime_sum}
    \end{subfigure}%
    \begin{subfigure}{.5\textwidth}
        \centering
        \includegraphics[width=0.85\linewidth]{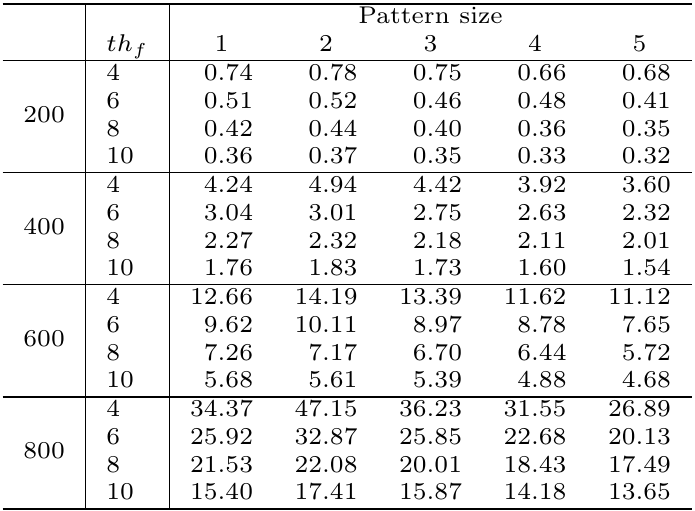}
        \caption{}
        \label{fig:npapers_vs_avgtime_agreement}
    \end{subfigure}
    \caption{Average running time (in seconds) for utility functions sum (a) and disagreement degree (b)}
    \label{fig:npapers_vs_avgtime}
\end{figure}

In a first series of experiments, we computed the average running times  for two different settings. The first one is a sort of a stress test: it involves all the facets provided by the use case and computes their sum via an external function call; thus, all input items and values are relevant for the computation and, consequently, a large number of patterns are expected. The second one is more realistic and computes the disagreement degree between one of the annotated sentiments and the decision about the corresponding paper, namely the percentage of pattern occurrences showing a positive sentiment on originality and a reject decision; in this case, given its simplicity, the utility function is directly encoded with ASP rules. Results for running times are shown in Figure \ref{fig:npapers_vs_avgtime} for increasing number of papers, pattern size, and occurrence threshold ($th_f$); each data point represents the average running time computed for 
utility thresholds equal to $[1,5,10,15,20,25,30]$ for the sum and $[15,50,75,100]$ for the disagreement degree. To rescale the dataset, we randomly selected the set of papers to be removed at each downsizing step, so that each instance is a strict subset of its larger versions. This selection process has been carried out once and the same datasets have been used throughout the experiments.

First of all, from the analysis of the results, it is possible to observe that the choice of the setting can significantly impact the performances. This is due to a combination of factors, including the amount of input data relevant to the calculation, the number of patterns satisfying the thresholds, and the complexity of the utility function itself.
In more detail, looking at Figure \ref{fig:npapers_vs_avgtime_sum}, we can observe that increasing the number of papers significantly increases execution time but, increasing the occurrence threshold significantly reduces required time. Moreover, especially for the bigger datasets, a higher occurrence threshold combined with a higher pattern size significantly reduces execution time. This behaviour can be mainly motivated by the portion of encoding for identifying frequent patterns; in particular, checking the number of occurrences within the choice rule, instead of using a standard constraint, makes rules more complex in general, but allows the evaluator to cut the search space more easily for bigger occurrence thresholds and, also, for larger patterns. In fact, it is worth pointing out that ground programs resulted to be smaller for increasing occurrence thresholds. We also observed that (results not shown due to space constraints)  the utility threshold does not significantly impact performances. Similar considerations can be drawn for the tests on  disagreement degree (Figure \ref{fig:npapers_vs_avgtime_agreement}). For these tests, it is worth observing the following differences with respect the previous ones using the sum: {\em (i)} input relevant facts are significantly lower (only words in sentences annotated with originality are useful); {\em (ii)}  required time and memory (see below) are significantly lower; {\em (iii)}   
the number of valid patterns is significantly lower and, in this case,  decreases with increasing utility thresholds. In particular, the maximum number of extracted patterns is 2480 with the sum, and 95 with the disagreement degree.

In a second series of experiments we considered the impact of external function calls in ASP. In particular, we consider two scenarios in which the utility function can be both expressed directly in ASP and in Python. 
Before going on with the analysis, let us emphasise once again that the constraints we are dealing with in this work to validate the patterns may come in the form of potentially complex functions, which are difficult, or even impossible, to express in ASP, and, as a result, external functions significantly broaden the scope of applicability of the proposed approach. Moreover, we observe that a deep analysis of when it is better to implement functions directly in ASP or in Python is out of the scope of this paper. The interested reader can find an extensive discussion on this aspect in \citep{Cuteri*17}.  

\begin{figure}
    \centering
    \begin{subfigure}{.5\textwidth}
        \centering
        \includegraphics[width=0.85\linewidth]{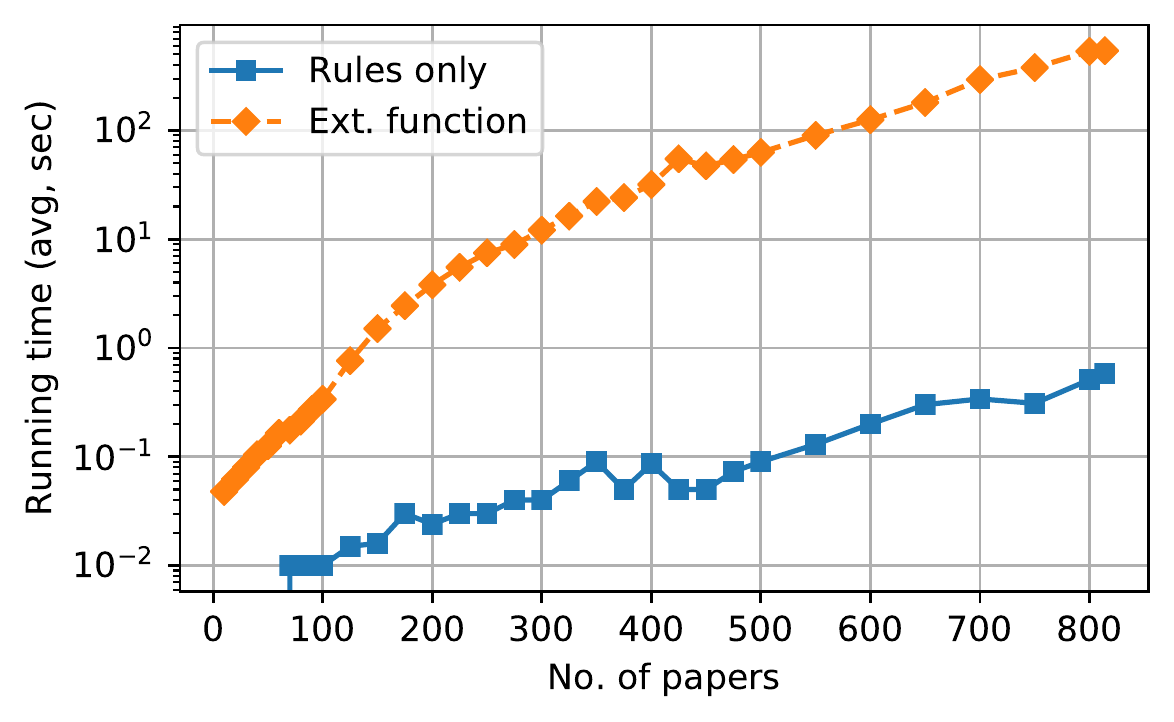}
        \caption{}
        \label{fig:external_vs_asp_sum}
    \end{subfigure}%
    \begin{subfigure}{.5\textwidth}
        \centering
        \includegraphics[width=0.85\linewidth]{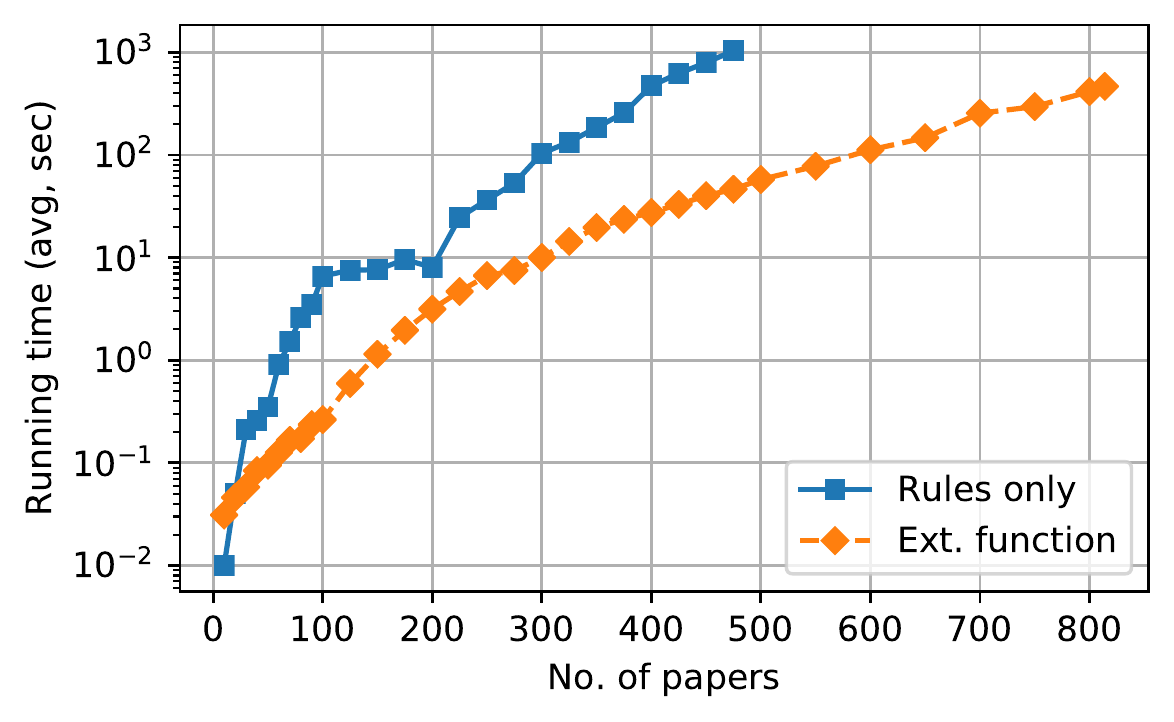}
        \caption{}
        \label{fig:external_vs_asp_product}
    \end{subfigure}
    \caption{Average running time (log scale) for utility functions sum (a) and product (b) using pure ASP or external functions}
    \label{fig:external_vs_asp}
\end{figure}

In the first experiment we considered the sum function, as introduced above, with the following parameters: pattern size 3, occurrence threshold 6, utility threshold 5. This is the worst case scenario for evaluating external functions, since ASP includes built-in functions for computing the sum and, in fact, results presented in Figure \ref{fig:external_vs_asp_sum} show that the impact of the external function calls may become consistent, especially when the number of papers increases. 

The second experiment considered the product function; in particular, for the purposes of this experiment, we added a facet to each review representing a (fictitious) probability, and we considered the combined probability as the utility function, i.e., the product of all the probabilities associated with pattern occurrences. Also in this case, we implemented a version of the program using external functions and a version implementing the product computation directly in ASP; it is worth observing that, in this case, no built-in function like the sum is available in ASP for the product and, thus, it must be computed with the support of recursive rules (probabilities have been obviously scaled to integers between 0 and 100). Parameters used in these tests are as follows: pattern size 3, occurrence threshold 6, utility threshold 5.
Results presented in Figure \ref{fig:external_vs_asp_product} clearly show that, in this case, the version with external functions significantly outperforms the one using ASP only, thus motivating the adoption of external sources of computation for utility functions which are not straightforward to be expressed in ASP.

\begin{figure}
    \centering
    \begin{subfigure}{.5\textwidth}
        \centering
        \includegraphics[width=0.85\linewidth]{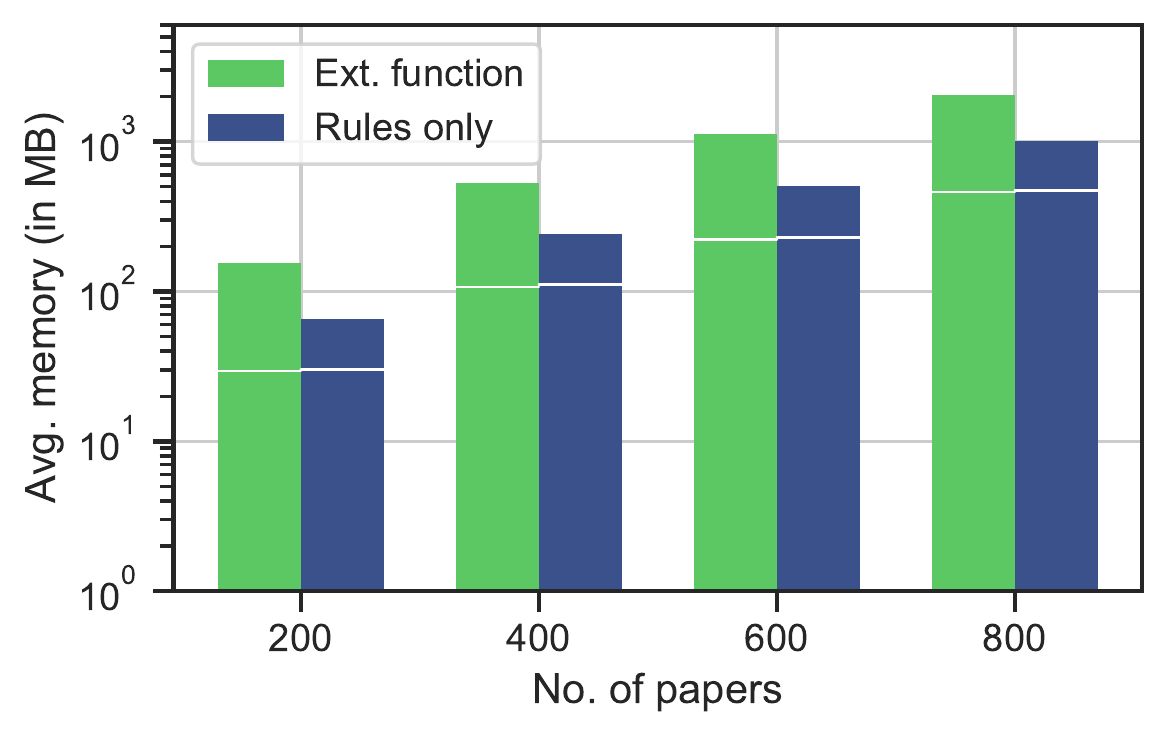}
        \caption{}
        \label{fig:memory_sum}
    \end{subfigure}%
    \begin{subfigure}{.5\textwidth}
        \centering
        \includegraphics[width=0.85\linewidth]{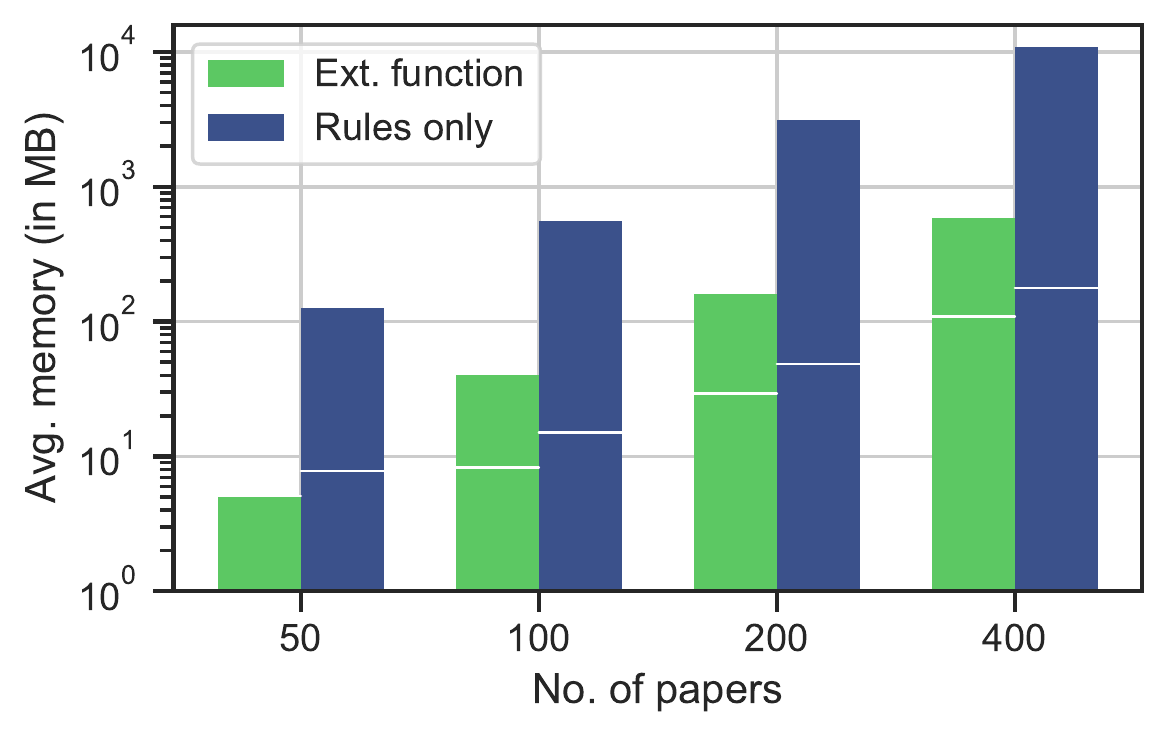}
        \caption{}
        \label{fig:memory_product}
    \end{subfigure}
    \caption{Average memory usage (log scale) for utility functions sum (a) and product (b). Horizontal white lines denote the boundary for the memory required by the grounder (bottom) and solver (top)}
    \label{fig:npapers_vs_mem}
\end{figure}

In order to show the impact of external functions on memory usage, we report in Figure \ref{fig:npapers_vs_mem} memory consumption for functions sum and product expressed directly in ASP or in Python. From the analysis of this figure, it is possible to observe, again, that for the sum function, the impact of external functions may become consistent. However, when we turn to the product function, the amount of memory required by the version using external functions is extremely lower. In both cases, the main difference is on the solver part, whereas the grounder part requires similar amounts of memory.

\subsection{Qualitative Analysis}
\label{sub:qualitative}

In this section we describe the results of some experiments carried out to show if, and how much, facets and utility functions introduced in Section \ref{sec:ehupm-framework} can help users in analyzing input data from different points of view and at different abstraction levels. 

In particular, we concentrate on the computation of coherence and disagreement degrees, introduced in Section~\ref{sub:pattern-utility-computation}, 
between the decision on a paper (Accept/Reject) and one of the eight aspects used to label review sentences~\citep{ChGoMu20}. In particular, the objective is to look for  patterns (sets of words) characterizing with good accuracy the relationship of interest and to qualitatively check if they are meaningful.

Before going into the details of the analysis, in order to better show the properties of our approach, consider that the paper review dataset contains 686 words with contrasting values between a ``sentiment'' (on one of the aspects among Appropriateness, Clarity, Originality, Empirical/Theoretical Soundness, Meaningful Comparison, Substance, Impact of Dataset/Software/Ideas and Recommendation) and the Decision. Among them, 481 different words are in sentences with an ``Originality'' sentiment; this situation leads to 929 patterns of size between 2 and 4 with at least 4 occurrences just related to ``Originality''. As a consequence, any classical approach based only on pattern frequency would be overwhelmed by a large amount of patterns.

Table \ref{tab:qualitative-coherence} shows some of the obtained results. Here, given the sentiment label on a sentence aspect $X$, the utility function measures the percentage of occurrences of the pattern showing coherence/disagreement between the sentiment on $X$ and the decision on the paper, as specified in the first column. The second column of Table \ref{tab:qualitative-coherence} reports the total number of patterns obtained by the approach for each test, whereas the third and fourth columns show an excerpt of the most relevant patterns along with their utilities. Parameters exploited for these tests are as follows: Minimum frequency=4, Minimum utility=70,\footnote{Observe that in this case the utility value expresses a percentage, as a consequence, 70 should be read as 70\% of the occurrences of the pattern satisfy the condition.} pattern size between 2 and 4.\footnote{Only for some specific cases we extended the size up to 6 in order to show more interesting patterns; these are annotated with a ``*''.}  It is worth pointing out that each run completed the computation in just few seconds, and switching between one test and the other involved few clicks and minor modifications to the ASP program.

\begin{table}[ht]
\centering
\caption{Qualitative analysis on coherence/disagreement degrees}
\label{tab:qualitative-coherence}
\tiny{
\begin{tabular}{lclc}
    \topline
    Setting & \# tot & Sample patterns & Utility 
   \midline
    \multirow{5}{5cm}{Originality(positive)-Reject} & \multirow{5}{*}{7} & paper interesting approach  & 100\\
    & & paper interesting write  & 80\\
    &  & simple nice idea   & 75\\
     &  & intriguing idea   & 75\\
      &  & idea proposed interesting    & 71
      \midline
     Originality(positive)-Accept &	1	& original approach &	75
     \midline
   \multirow{4}{5cm}{Originality (negative)-Accept} & \multirow{4}{*}{6} & incremental paper & 80\\
    & & easy conference paper &	75 \\
     & & easy follow interesting &	75 \\
     & & paper write easy read & 	75
     \midline
   \multirow{6}{5cm}{Disagreement between Clarity and Decision: Clarity(positive)-Reject or Clarity(negative)-Accept} & \multirow{6}{*}{71} & easy paper overall  & 100\\
    & & easy conference paper &	100 \\
    & & easy follow interesting &	85 \\
    & & paper write easy read &	85 \\
    & & presentation generally easy  & \\
    & & follow interesting result$^*$ &	80
    \midline
    Appropriateness(positive)-Accept &	0	&  &
    \midline
  \multirow{2}{5cm}{Appropriateness(negative)-Accept} & \multirow{2}{*}{2} & niche audience  & 75\\
    & & paper accessible &	75
    \midline
  \multirow{4}{5cm}{Impact of Dataset/Software/Ideas(positive)-Accept} & \multirow{4}{*}{20} & paper write clearly present  & 100\\
    & & dataset significant &	100 \\
    & & proposed method datasets baselines &	75 \\
    & & overall contribution &	74
   \midline
  \multirow{5}{5cm}{Recommendation(negative)-Reject} & \multirow{5}{*}{84} & paper reject  & 100\\
    & & accept conference need significant &	75 \\
     & & acceptance work need significant &	75 \\
     & & recommend acceptance conference  &	 \\ 
     & & need significant work$^*$ & 75
     \midline
 Recommendation(positive)-Reject &	0	&  &
 \midline
 Rating low (1-4)-Accept &	0	&  & \midline
 \multirow{5}{5cm}{Rating high (5-10)-Accept} & \multirow{5}{*}{173} & paper write clearly easy   & 100\\
    & & paper follow write clearly &	100 \\
    & & dataset new &	100 \\
    & & theoretical approach &	100 \\
    & & present clearly &	75  
    \botline
\end{tabular}
}
\end{table}

From the analysis of Table \ref{tab:qualitative-coherence} we can draw the following observations: \emph{(i)} The number of valid patterns characterizing each setting is very low; this allows the user to concentrate on the most relevant ones. \emph{(ii)} Patterns extracted in the different settings are quite different; this indicates a good specificity of derived patterns.  \emph{(iii)} For some settings, the number of derived patterns is 0; this means that the corresponding situation cannot be characterized. As far as this last observation is concerned, consider the very interesting situation analyzing Recommendation and Reject: in this case no patterns characterize a positive recommendation (which we recall is a derived sentiment from the sentences) associated with a reject decision, whereas 84 patterns characterize the opposite situation, namely negative recommendation and reject; this provides an interesting insight on the appropriateness of derived patterns. We also carried out a similar analysis relating the Rating facet, associated with the Review (which is an object in our framework), and the Decision facet, associated with the Paper (which is a container in our framework). Results are shown in the last two rows of Table \ref{tab:qualitative-coherence}; these confirm the analysis relating recommendation  and decision and show how easy is to switch the analysis also between different levels of abstraction.

\section{Related Work}
\label{sec:related-work}

Considering its importance as a research topic in the data mining field, HUPM received a huge interest both from academic community and industrial practitioners. HUPM finds application both in classical contexts where (normal) itemset mining was born, such as  market analysis, and in novel ones such as biomedicine, mobile computing, etc. \citep{SuSr19}. Thanks to its exhibited versatility, a huge volume of research studies has been produced regarding HUPM and its variants. According to Semantic Scholar\footnote{\url{https://www.semanticscholar.org}}, in the last twenty years more than 8,000 studies related to HUPM have been published.

In order to better characterize the work related to our approach and to discern among the plethora of presented studies, we divide the analysis of the related literature in two parts. The former provides a bird's-eye view of the general literature about HUPM, focusing on variations of the utility measure; the latter focuses on some declarative-based approaches to pattern mining.

It is worth pointing out that, to the best of our knowledge, no previous work addressed in a comprehensive way the various extensions introduced in the present paper; additionally, our work is the first ASP-based solution to the e-HUPM problem. 

\subsection{Approaches for HUPM and its variants}
As previously pointed out, HUPM is a general container for several high utility-based mining approaches~\citep{Fou-Vi*19,Gan*19a}. The corresponding research studies can be divided in several groups, depending on the aspect one is interested to analyze, such as employed algorithms or pruning strategies~\citep{Krishnamoorthy17,YuRyLeFu17,WuLiTa19,YuNaLeYo19,SuSu20}. In our context, we mainly focus on research  that studies  utility measures and their variants~\citep{YaHaGe06,ShZhYa02,Fournier*14,Cagliero*14,Fournier*20a,Deng20}.

One of the first works on unifying utility based measures is presented in~\citep{YaHaGe06}, where the authors introduce a unified utility framework in which a user defined function $f(x,y)$ is exploited to define utilities. Here, the significance of an item is measured by two parts: one is the statistical significance of the item, measured by $x$, and is an objective measure; the other is the semantic significance of the item, measured by $y$, which is a subjective measure dependent on the user perception of utility. Moreover, a weight to each transaction, called vertical weight, can be assigned. The authors show that this generalization can assign semantic significance at different levels of granularity (item, transaction, and cell level). 
The approach proposed in~\citep{YaHaGe06} and our own share some similarities, but the present work further extends the pattern mining capabilities, as specified next. \emph{(i)} First of all, \citep{YaHaGe06} attempts to provide different levels of abstraction for the identification of pattern utility; the present paper has a similar goal but extends it not only to transactions but also to generic abstraction layers, enabling the introduction of more advanced utility functions. \emph{(ii)} Allowing $f$ to be a user defined function extends the scope of application; however, only one property at a time can be used to state the semantic significance of an item; in our approach, the introduction of facets allows not only to consider several properties at the same time, but also to combine them dynamically. \emph{(iii)} The introduction of facets also on transactions, objects, and containers allows not only analyses at different levels of granularity, as done in~\citep{YaHaGe06}, but also the identification of patterns expressing correlations (and possibly causality) between different properties, at different levels. \emph{(iv)} Finally, we have shown that beside classical sum and product operations on utility values, exploited in ~\citep{YaHaGe06}, more complex functions like Pearson correlation or Multiple correlation, allowed by the availability of facets, broaden the scope of potential analyses.

An effort to define a general model, called Objective-Oriented Utility-Based Association mining, in which mined association patterns  are both statistically and semantically related to a user's given objective, is explained in~\citep{ShZhYa02}. 
Here, the database is composed of a set of records structured in a pre-defined set of attributes, and an association rule is enhanced with an objective and  utility value. Thus, the approach aims at deriving patterns from the records that both statistically and semantically support a given objective, by taking into account minimum support, confidence and utility. While the scope of~\citep{ShZhYa02} and our own are not completely overlapping, namely association rule mining vs high utility pattern mining, our approach can embrace the concept of objective-driven mining. 
Thus, with the adoption of suitable pattern masks, facets, and utility functions, our proposal can be considered to some extent as an extension of the concepts presented in~\citep{ShZhYa02}.

The authors in~\citep{Deng20} define a novel measure called occupancy and formulate the problem of mining high occupancy itemsets. The occupancy is defined as the sum of the ratio between the cardinality of an itemset $p$ and the cardinality of the transactions that contain $p$. Then, a high occupancy itemset is an itemset whose occupancy is not below a given threshold. 
Interestingly, the approach described in~\citep{Deng20} can be seen as a special case of our approach, where a \emph{horizontal first} utility function class is applied.

It is also worth pointing out that different extensions of the HUPM problem are presented in the literature. Different aspects are considered, and each aspect aims at addressing the limitations of the classical problem~\citep{Fou-Vi*19}. As an example, an HUPM algorithm may show a large number of patterns to the user if the minimum utility threshold given in input is too low, thus it can be very hard for a user to analyze and describe retrieved patterns. To this end, several concise representations for high utility patterns have been studied. Four main representations are: closed high utility patterns, maximal patterns, generators of high utility patterns and maximal high utility patterns~\citep{Fou-Vi*19}; moreover, in~\citep{Gan*19b} the Kulc measure has been adopted to extract non-redundant strongly correlated and profitable patterns. In our approach, the number of retrieved patterns can be reduced by defining more stringent utility functions; however, we can see our approach and the ones mentioned above as orthogonal, since they do not focus on the definition of utility, but rather on the representation and filtering of obtained results.  

Other different aspects are studied in the literature, such as HUPM with negative utility and HUPM with discount strategies~\citep{Fou-Vi*19}. These  are all covered by our framework, which can be seen as an extension of these special cases with further potentialities. 

In addition to the aforementioned approaches, it is worth noting that HUPM has been employed in several different contexts, such as topic detection~\citep{ChPa19}, relevant information extraction~\citep{Belhadi*20}, aspect-based sentiment analysis~\citep{Demir*19}, and mining in noisy databases~\citep{Baek*20}. This variety of application contexts for HUPM further motivates our framework which, thanks to its generality, can accommodate very different settings, as it has been shown through the paper.

Finally, there are also different pattern extraction methods, not related to HUPM, which could be indirectly encompassed by our framework. An example is that of subgroup discovery~\citep{Herrera*11,Atzmueller15}. Briefly, subgroup discovery is a method for knowledge discovery in databases whose aim is to identify relations between a target variable and many explaining and independent variables, determined by a quality function that can be flexibly defined~\citep{Atzmueller15}. Intuitively, it means to discover the subgroups of the population that are statistically most interesting and, therefore, can be identified as a sort of pattern extraction technique. This task has been addressed with several approaches, including those exploiting evolutionary algorithms~\citep{delJesus*07}, complex network aspects~\citep{Skrlj*18} and inductive logic programming~\citep{CeZe12}. Although our notion of utility could encompass similar quality measures as those used in subgroup discovery, the aim of our framework is  different from the task of subgroup discovery. Indeed, this last task could be further investigated in a future work by considering the multi-level structure offered by our approach. The application of our framework we presented in Section~\ref{sec:application} differs from subgroup discovery in terms of the purpose of the knowledge discovery task. The goal of classification in our application is to generate a model for each class that contains rules representing class characteristics regarding the descriptions of training examples. This model is used to predict the class of unknown objects. In contrast, subgroup discovery attempts to discover interesting relations with respect to the property of interest \citep{Helal16}.

\subsection{Declarative-based approaches for pattern mining}

The usage of declarative systems to tackle combinatorial problems is a consolidated approach which attracted a peculiar interest from researchers. The possibility of coupling the power of a high-level declaration with an optimised solver allows users to specify how patterns should be and not how they should be computed. There are different existing approaches that blend together the expressiveness and readiness to use of a declarative system within the problem of pattern mining~\citep{Jarvisalo11,Gebser*16,SaGuNe17,PaStMi19,GuMoQu14,GuMoQuSc16}. The main objective of this section is to show the growing interest in  declarative-based approaches for pattern mining as an alternative to dedicated algorithms, especially when the main focus is to add expressiveness to the standard problem at hand. 
To the best of our knowledge there is no ASP-based approach tackling the HUPM problem, even in its basic form; then, with respect to this aspect, our approach moves a step forward in the application of ASP with external functions in interesting data mining contexts.

The seminal work in~\citep{Jarvisalo11} exploits ASP in the task of finding all frequent itemsets: each itemset is an answer set, thus the enumeration of all answer sets corresponds to the enumeration of all frequent itemsets. Although both our approach and the one in~\citep{Jarvisalo11} exploit ASP, the latter addresses only the standard frequent itemset setting; in our extended HUPM setting, we also exploit the recent ASP language extensions for the application of complex functions on sets of data inferred during the evaluation of the program.

In~\citep{Gebser*16}, the context of sequence mining is addressed, and an ASP-based mining approach is presented. Here, the focus is to express and incorporate knowledge in the mining process. The authors consider  frequent (closed or maximal) sequential patterns and complex preferences on patterns. Frequent patterns are mined following similar encoding principles of~\citep{Jarvisalo11}, whereas preference-based mining is accomplished by exploiting the \emph{asprin} system~\citep{BrDeRoSc15} which allows expressing combinations of qualitative and quantitative preferences among answer sets. 
Rare sequential pattern mining is considered in~\citep{SaGuNe17}.

A general and hybrid approach to pattern mining is presented in~\citep{PaStMi19}, where different dedicated mining systems are employed to detect frequent patterns and  ASP is used to filter the results. Here, the tasks considered are itemset, sequence and graph mining, where local (frequency, size, and cost) and global (condensed representations such as maximal, closed, and skyline) constraints are combined in a generic way. The declarative approach here is devoted to post-process the patterns in order to select the valid ones.

Mining serial patterns, i.e., frequent sequential patterns from a unique sequence of itemsets, is introduced in~\citep{GuMoQu14}, whereas sequential pattern mining with two representations of embeddings (fill-gaps and skip-gaps) and several kinds of patterns are considered in~\citep{GuMoQuSc16}.

Other declarative approaches, not constrained to ASP and based, e.g., on constraint programming or SAT, are also discussed in the literature~\citep{NeDrGuNi13,Guns*17,GuPaNe16}. 
Itemset mining is considered in~\citep{NeDrGuNi13}, in which a novel algebra for programming pattern mining problems is introduced. It allows for the generic combination of constraints on individual patterns with dominance relations between patterns. Dominance relations are pairwise preferences between patterns, used to express the idea that a pattern $p_1$ is preferred over another pattern $p_2$. Various settings are described with combinations of constraints and dominance relations, including maximal and closed itemset mining, sky patterns, etc. Dominance programming offers a great extensibility, although it is not as general as our approach. Furthermore, the context considered here only aims at itemset mining. Itemset, sequence and graph mining tasks are also considered in a framework introduced in~\citep{GuPaNe16}, in which these pattern types are studied under the lens of constraint-based mining~\citep{MaTo97}. A declarative framework for constraint-based mining is presented in~\citep{Guns*17}, where the objective is to bridging the methodological gap between data mining and constraint programming by providing a framework to support constraint-based pattern mining. In particular, the authors offer several contributions, spanning from a novel library of functions and constraints in the MiniZinc language to support modeling itemset mining tasks, to the automatic composition of execution strategies involving different solving methods. Indeed, the work in~\citep{Guns*17} shares some common similarities with our proposed approach. In particular, both the approaches provide a general framework for high-utility mining, and they are based on declarative paradigms. Our approach supports the various extensions to HUPM introduced in Section \ref{sec:ehupm-framework}, which are not supported by~\citep{Guns*17}. Nevertheless, \citep{Guns*17} also deals with multiple databases, which is a feature not currently supported by our approach.

An interesting task studied in the literature is the discovery of skyline patterns, or ``skypatterns''~\citep{Soulet*11, Ugarte*17}. Skypatterns draw their definition from the economics and multi-criteria decision making worlds, in which they are commonly called Pareto efficiency or optimality queries. Intuitively, given a multidimensional space where a preference is defined for each dimension, a point $a$ dominates another point $b$ if $a$ is better than $b$ in at least one dimension, and $a$ is not worse than $b$ on every other dimension. Given a set of patterns, the skyline set contains patterns that are not dominated by any other pattern~\citep{Ugarte*17}. Skypattern mining provides an easy expression of preferences over patterns by combining several measures on patterns. An approach for skypattern mining is presented in~\citep{Ugarte*17}. Here, the authors address the discovery of skypatterns by proposing a unified methodology based on the following contributions, namely {\em (i)} a condensed representation of patterns, {\em (ii)} a dynamic method based on constraint programming that allows a continuous refinement of the skyline constraints, and {\em (iii)} the notion of particular local extrema in the search space of the problem. The work in~\citep{Ugarte*17} shares few similarities with ours; actually the two approaches can be considered complementary.

\section{Conclusion}
\label{sec:conclusion}

In this paper we introduced a general framework for HUPM with several extensions that significantly broaden the applicability of HUPM even in non classical contexts. We provided an ASP based computation method, which exploits the most recent extensions of ASP, and we have shown that this solution allows for a reasonable and versatile implementation. 
A real use case  on paper reviews has been employed as a fil-rouge to analyze the different aspects of the proposed framework; in particular, we have shown that the introduction of facets and suitable advanced utility functions can both reduce the amount of relevant patterns and provide deep insights on the data, thus advancing the state-of-the-art on the interesting topic of high utility pattern mining. An application to a biomedical context has also been presented and tested, which showed how the novel pattern mining framework can be suitably exploited to design effective prediction models on biomedical data. 

The presented work is just a first step in this new direction of utility-based analyses and several future research directions are now open. First of all, it will be interesting to apply the new features of the framework to a wide spectrum of contexts, such as biomedical data analysis, IoT data analysis, and fraud detection. A classification of the computational properties of the various extended utility functions will be also useful to devise ad-hoc algorithms. In particular, we plan to develop dedicated and efficient algorithms for specific settings of interest, in order to provide fast computational methods for extended high utility patterns. Another interesting line of future research regards a deeper study of the application of our e-HUPM to prediction tasks in various contexts.

\paragraph*{Competing interests} The authors declare none.

\bibliographystyle{tlplike}

\end{document}